\documentclass[review]{elsarticle}

\usepackage{lineno,hyperref}
\modulolinenumbers[5]

\journal{ISPRS Open Journal of Photogr. and Rem. Sens.}

\biboptions{authoryear}

\usepackage{units}

\usepackage{booktabs}  
\usepackage{multirow}

\usepackage{epstopdf}
\AppendGraphicsExtensions{.tif}
\usepackage{color}
\usepackage{colortbl}
\usepackage{subfigure}
\usepackage{adjustbox}
\usepackage{lscape}
\usepackage{rotating}

\usepackage[acronym,toc,shortcuts]{glossaries}

\newacronym{ALS}{ALS}{Airborne Laser Scanning}
\newacronym{CNN}{CNN}{Convolutional Neural Network}
\newacronym[plural=COGs, firstplural=Centers of Gravity (COGs)]{COG}{COG}{Center of Gravity}
\newacronym{DL}{DL}{Deep Learning}
\newacronym{GSD}{GSD}{Ground Sampling Distance}
\newacronym{GT}{GT}{Ground Truth}
\newacronym{H3D}{H3D}{Hessigheim~3D}
\newacronym[plural=MBBs, firstplural=Minimum Bounding Boxes (MBBs)]{MBB}{MBB}{Minimum Bounding Box}
\newacronym{ML}{ML}{Machine Learning}
\newacronym{MVS}{MVS}{Multi-View Stereo}
\newacronym{OA}{OA}{Overall Accuracy}
\newacronym[plural=PCs]{PC}{PC}{Point Cloud}
\newacronym{RF}{RF}{Random Forest}
\newacronym{UAV}{UAV}{Unmanned Airborne Vehicle}
\newacronym{V3D}{V3D}{Vaihingen~3D}

\definecolor{LowVeg}{RGB}{178,203,47}
\definecolor{ISurf}{RGB}{183,178,170}
\definecolor{Vehicle}{RGB}{32,152,163}
\definecolor{UFurn}{RGB}{168,33,107}
\definecolor{Roof}{RGB}{255,122,89}
\definecolor{Facade}{RGB}{255,215,136}
\definecolor{Shrub}{RGB}{89,125,53}
\definecolor{Tree}{RGB}{0,128,65}
\definecolor{Soil}{RGB}{170,85,0}
\definecolor{VertSurf}{RGB}{252,225,5}
\definecolor{Chimney}{RGB}{128,0,0}

\begin{document}

\begin{frontmatter}

\title{The Hessigheim 3D (H3D) Benchmark on Semantic Segmentation of High-Resolution 3D~Point~Clouds and Textured Meshes from UAV LiDAR and Multi-View-Stereo}

\author[ifp]{Michael Kölle\corref{mycorrespondingauthor}} 
\ead{michael.koelle@ifp.uni-stuttgart.de}
\author[ifp]{Dominik Laupheimer\corref{mycorrespondingauthor}} 
\ead{dominik.laupheimer@ifp.uni-stuttgart.de}
\author[ifp]{Stefan Schmohl} 
\author[ifp]{Norbert Haala}
\address[ifp]{Institute for Photogrammetry, University of Stuttgart, Germany}

\author[ipi]{Franz Rottensteiner}
\address[ipi]{Institute of Photogrammetry and GeoInformation, Leibniz University Hannover, Germany}
\author[ethz]{Jan Dirk Wegner}
\address[ethz]{EcoVision Lab, University of Zurich \& ETH Zurich, Switzerland}
\author[delft]{Hugo Ledoux}
\address[delft]{Faculty of the Built Environment \& Architecture, Delft University of Technology, the Netherlands}

\cortext[mycorrespondingauthor]{Corresponding authors}

\begin{abstract}
Automated semantic segmentation and object detection are of great importance in geospatial data analysis. However, supervised machine learning systems such as convolutional neural networks require large corpora of annotated training data. Especially in the geospatial domain, such datasets are quite scarce. Within this paper, we aim to alleviate this issue by introducing a new annotated 3D dataset that is unique in three ways: i)~The dataset consists of both an Unmanned Aerial Vehicle (UAV) laser scanning point cloud and a 3D textured mesh. ii)~The point cloud features a mean point density of about \unit[800]{pts/m$^2$} and the oblique imagery used for 3D mesh texturing realizes a ground sampling distance of about \unit[2-3]{cm}. This enables the identification of fine-grained structures and represents the state of the art in UAV-based mapping. iii)~Both data modalities will be published for a total of three epochs allowing applications such as change detection. The dataset depicts the village of Hessigheim (Germany), henceforth referred to as H3D. It is designed to promote research in the field of 3D data analysis on one hand and to evaluate and rank existing and emerging approaches for semantic segmentation of both data modalities on the other hand. Ultimately, we hope that H3D will become a widely used benchmark dataset in company with the well-established ISPRS Vaihingen 3D Semantic Labeling Challenge benchmark (V3D). The dataset can be downloaded from \url{https://ifpwww.ifp.uni-stuttgart.de/benchmark/hessigheim/default.aspx}.
\end{abstract}


\end{frontmatter}


\glsresetall
\section{Introduction}\label{sec:INTRODUCTION}

Supervised Machine Learning (ML), especially embodied by Convolutional Neural Networks (CNNs), has become state of the art for automatic interpretation of various data. However, the applicability and acceptance of such approaches are greatly hindered by the lack of labeled datasets for both training and evaluation (and, consequently, for the verification of their quality). For that purpose, large datasets of labeled 2D imagery were established, for example the \textit{ImageNet} dataset \citep{Deng2009}. As such an extensive annotation process cannot be accomplished by a single person or group, crowdsourcing was employed. Whereas 2D imagery can be very well interpreted by non-experts (i.e., crowdworkers), labeling 3D data is much more demanding. Although first investigations were conducted on employing crowdworkers for 3D data annotation \citep{Dai2017,Herfort2018,Walter2020,Koelle2020}, these approaches typically try to avoid deriving a full pointwise annotation. This is achieved either by working on object level or by focusing only on necessary points by exploiting active learning techniques. However, at least for evaluating ML models for semantic segmentation, full annotations are beneficial, which are typically acquired by experts. In case of 3D data, existing datasets can be categorized into three different domains (comprehensive literature reviews are given by \cite{Griffiths2019} and \cite{Xie2020}): indoor data, outdoor terrestrial data, outdoor airborne data. \\
\begin{figure}[htbp]
\begin{center}
		\includegraphics[width=0.65\columnwidth,trim=0 10 120 10, clip]{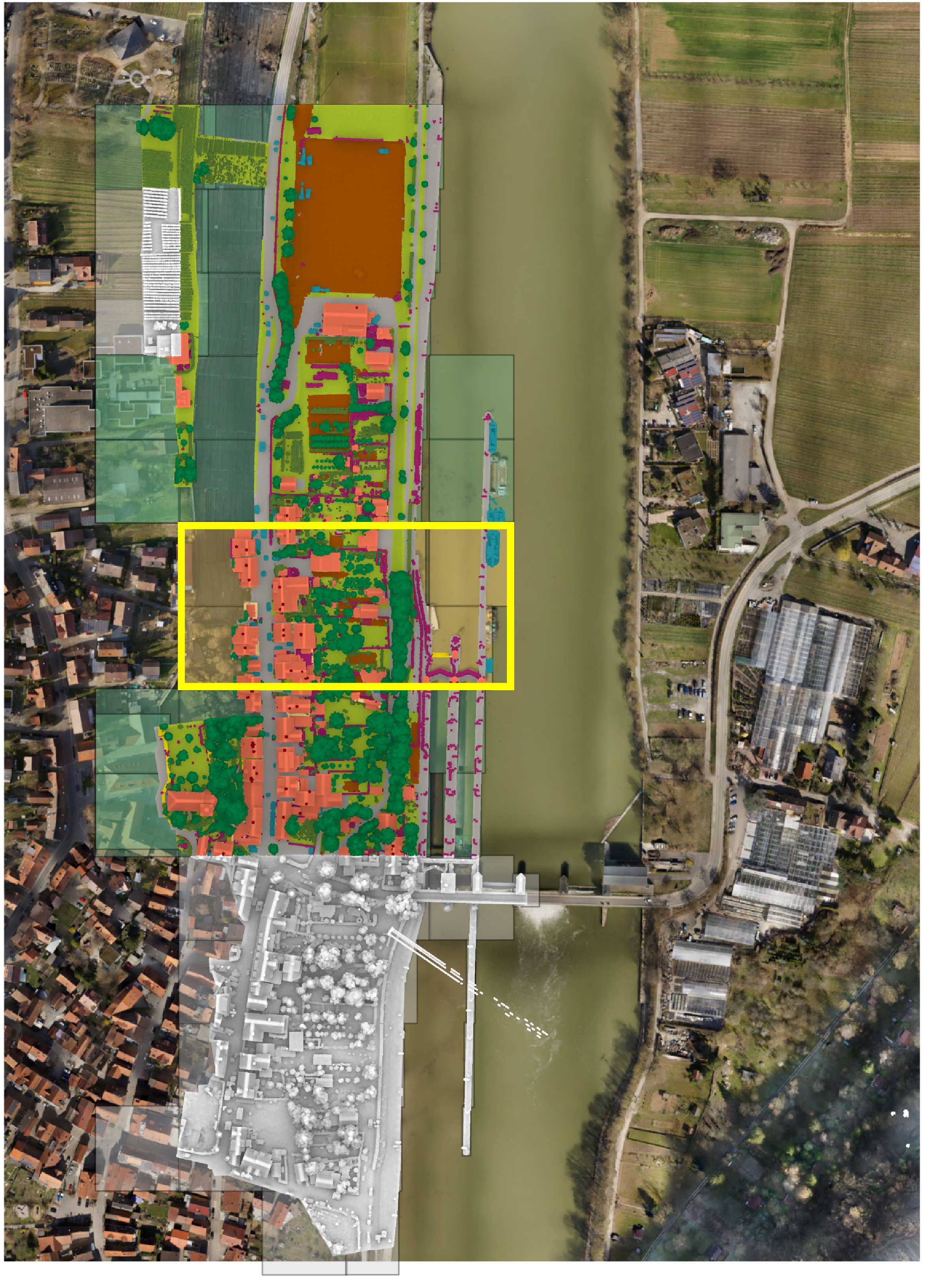}
	\caption{Partition of the H3D(PC) dataset into training (\textit{class colors}; see Table~\ref{tab:classes}), validation (\textit{yellow box)} and test set (\textit{grey}). Data splits of H3D(Mesh) are identical but organized in tiles (\textit{individually colored according to data splits in transparent manner}). As can be seen, H3D(Mesh) exceeds H3D(PC) in terms of covered area. }
\label{fig:Splits}
\end{center}
\end{figure}

\textbf{i) Indoor data.} Indoor 3D data typically depicts various scenes in living space or working environments, often captured by RGB-D sensors such as \textit{Microsoft Kinect} \citep{Silberman2012,Song2015}. Generally, even non-experts are familiar with such scenes, which is why the annotation process can be outsourced to crowdworkers, for instance via \textit{Amazon Mechanical Turk} \citep{Buhrmester2011}. This is realized by providing (pre-segmented) data embedded in easy to handle tools as described in \cite{Dai2017} and \cite{Silberman2012}. For better interpretability, often multi-modality is realized in the sense that acquired point clouds are meshed to obtain a well-defined closed surface representation \citep{Hua2016}.

\textbf{ii) Outdoor terrestrial data.} Capturing outdoor terrestrial data has become most popular in the context of autonomous driving. Cars destined for self-driving are equipped with a great variety of different sensors such as cameras, laser scanners, and odometers. Often only the combination of these sensors allows a comprehensive understanding of the complete scene, which is studied extensively on the basis of the well-known KITTI dataset \citep{Geiger2012}. Although Mobile Laser Scanning (MLS) point clouds of typical urban scenes are often provided as stand-alone products \citep{Roynard2017, Munoz2009, Hackel2017}, the concurrent availability of LiDAR data and imagery in the form of meshes is often pursued \citep{Riemenschneider2014}. \cite{Caesar2020} provide a unique multi-modal dataset for autonomous driving applications by the combination of both cameras and ranging sensors (i.e., LiDAR \& RADAR) for 3D object detection.

\textbf{iii) Outdoor airborne data.} Datasets of this category are often referred to as (large-scale) geospatial data and deviate from the previous ones due to a significantly increased distance between target and sensor, which is attached to an airborne platform (mostly small aircraft). So far, publicly available datasets provide labeled point clouds obtained from a single sensor, either a camera \citep{Hu2020} or a LiDAR sensor \citep{Varney2020}. One prominent example of the latter case is the Vaihingen 3D (V3D) dataset acquired by \cite{Cramer2010}, which served as the basis for the ISPRS 3D Semantic Labeling benchmark \citep{Niemeyer2014}. Additionally, some national mapping agencies publish large-scale annotated LiDAR point clouds as open data, which typically realize a rather coarse class catalog and a point density of up to about \unit[40]{pts/$\text{m}^2$} \citep{ANH3,FinlandNS}. For obtaining higher LiDAR point densities, the carrier of the sensor can be replaced by a helicopter platform allowing the generation of data with point densities of up to about \unit[350]{pts/$\text{m}^2$} \citep{Zolanvari2019}. Due to the high point density and the resulting depiction of fine structures, the authors opt for expanding the corresponding class catalog. \\

The Hessigheim 3D (H3D) dataset presented in this paper belongs to the third group but differs from other datasets because it is the first ultra-high resolution, fully annotated 3D dataset acquired from a LiDAR system and cameras integrated on the same Unmanned Aerial Vehicle (UAV) platform. This results in a unique multi-modal scene description by a LiDAR point cloud H3D(PC) and a textured 3D mesh H3D(Mesh). Hence, properties unique for these two acquisition methods can be efficiently combined, which offers new possibilities for high-accuracy georeferencing \citep{Glira2019} and semantic segmentation \citep{Laupheimer2020}. We consider H3D to be the logical successor of V3D, which was already captured in 2008 and therefore no longer represents the state of the art. As H3D was acquired from a UAV platform by the usage of state-of-the-art sensors, a point density that is about a hundred times higher compared to V3D can be achieved, allowing the expansion of the class catalog of V3D.

The rest of this paper focuses on presenting H3D as a new benchmark dataset. This includes a detailed presentation of data acquisition, the registration process, and a discussion of the unique characteristics of H3D (Sections~\ref{sec:datacapt}-\ref{sec:Mesh}). Sections~\ref{sec:catalog} and~\ref{sec:GT} are dedicated to present the class catalog and the annotation process for H3D(PC) and H3D(Mesh). As we aim at generating a benchmark for semantic segmentation, Section~\ref{sec:splits} describes the general structure of H3D, i.e., the partitioning into disjoint subsets for training, validation, and testing. As labels of the test set are not disclosed to the public, labels are to be predicted by participants (Section~\ref{sec:task}) and transmitted to the authors for evaluation (Section~\ref{sec:metrics}). We present first results based on two state-of-the-art approaches for semantic segmentation to kick off the benchmark process and to give first baseline results (Section~\ref{sec:baselines}) before concluding with a summary in Section~\ref{sec:conclusion}.

\section{The H3D Dataset}\label{sec:The H3D Dataset}
The H3D dataset was originally captured in a joint project of the University of Stuttgart and the German Federal Institute of Hydrology (BfG) for detecting ground subsidence in the sub-mm accuracy range. For this monitoring application, the area of interest, which is the village of Hessigheim, Germany (see Figure~\ref{fig:Splits}), was surveyed at multiple epochs in March 2018, November 2018, and March 2019. Although all those datasets will be made publicly available, this paper only covers the first epoch (i.e., March 2018) as only this dataset has been labeled so far. Concerning later epochs, the data publication shall follow in the course of 2021.

\subsection{Capturing H3D}\label{sec:datacapt}
\begin{figure}[htbp]
\begin{center}
	\includegraphics[width=0.8\columnwidth,trim=0 0 0 55, clip]{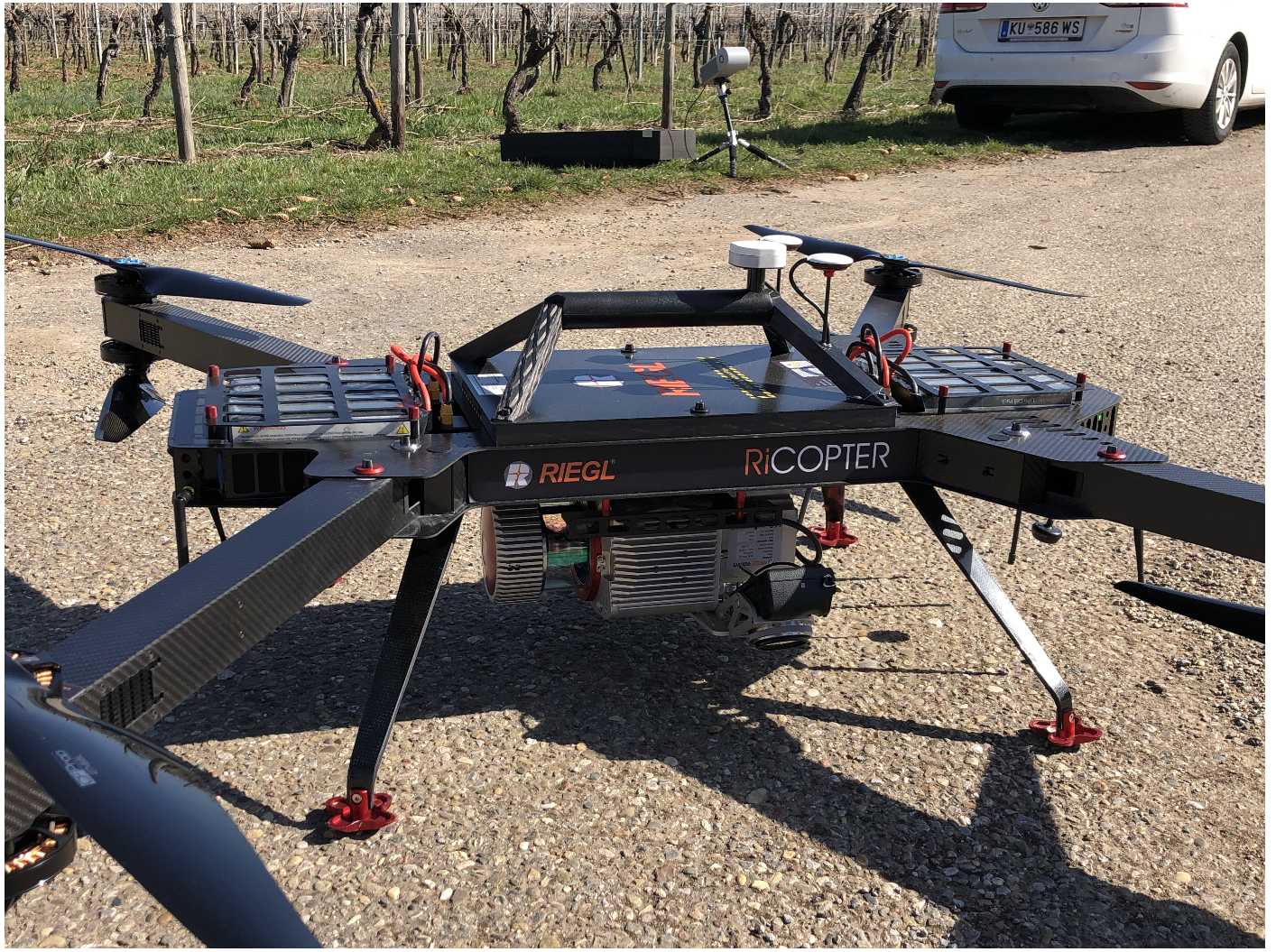}
	\caption{\textit{Ricopter} platform equipped with \textit{Riegl VUX-1LR} scanner and two oblique \textit{Sony Alpha 6000} cameras used for capturing H3D.}
    \label{fig:Ricopter}
\end{center}
\end{figure}

In all three epochs, our sensor setup was constituted of a \textit{RIEGL VUX-1LR} scanner and two oblique \textit{Sony Alpha 6000} cameras integrated on a \textit{RIEGL Ricopter} platform (see Figure~\ref{fig:Ricopter}). Considering a height above ground of \unit[50]{m}, we achieved a laser footprint of less than \unit[3]{cm} and a ground sampling distance of \unit[2-3]{cm} for the cameras. Using this setup, we obtain two distinct data representations: i) H3D(PC) and ii) H3D(Mesh) (see Section~\ref{sec:PC} and Section~\ref{sec:Mesh} respectively).

\subsection{H3D(PC)}\label{sec:PC}
\begin{figure*}[htbp]
\begin{center}
		\includegraphics[width=0.55\columnwidth,trim=0 0 0 0, clip]{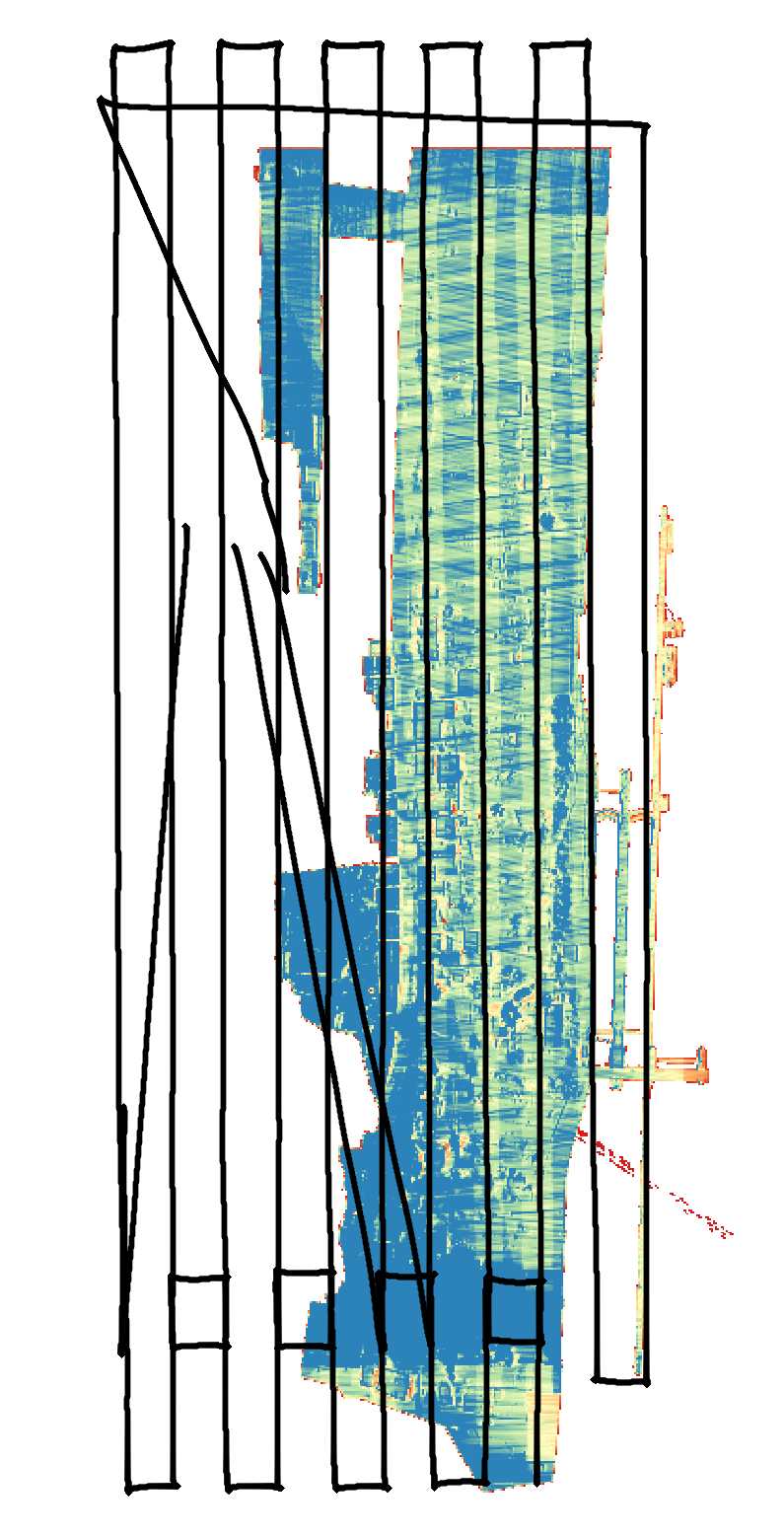}
		\hspace{0.5cm}
		\raisebox{0.5\height}{\includegraphics[width=0.18\columnwidth,trim=0 0 0 0, clip]{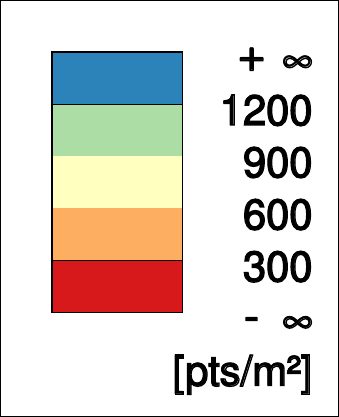}}
	\caption{Achieved point density of H3D(PC). Different point densities are due to diagonal strips for further block stabilization beneficial for adjustment of LiDAR strips. Flight trajectories of LiDAR strips are shown in \textit{black}.}
    \label{fig:Density}
\end{center}
\end{figure*}

\begin{figure}[htbp]
\begin{center}
		\includegraphics[width=0.93\columnwidth,trim=20 20 20 20, clip]{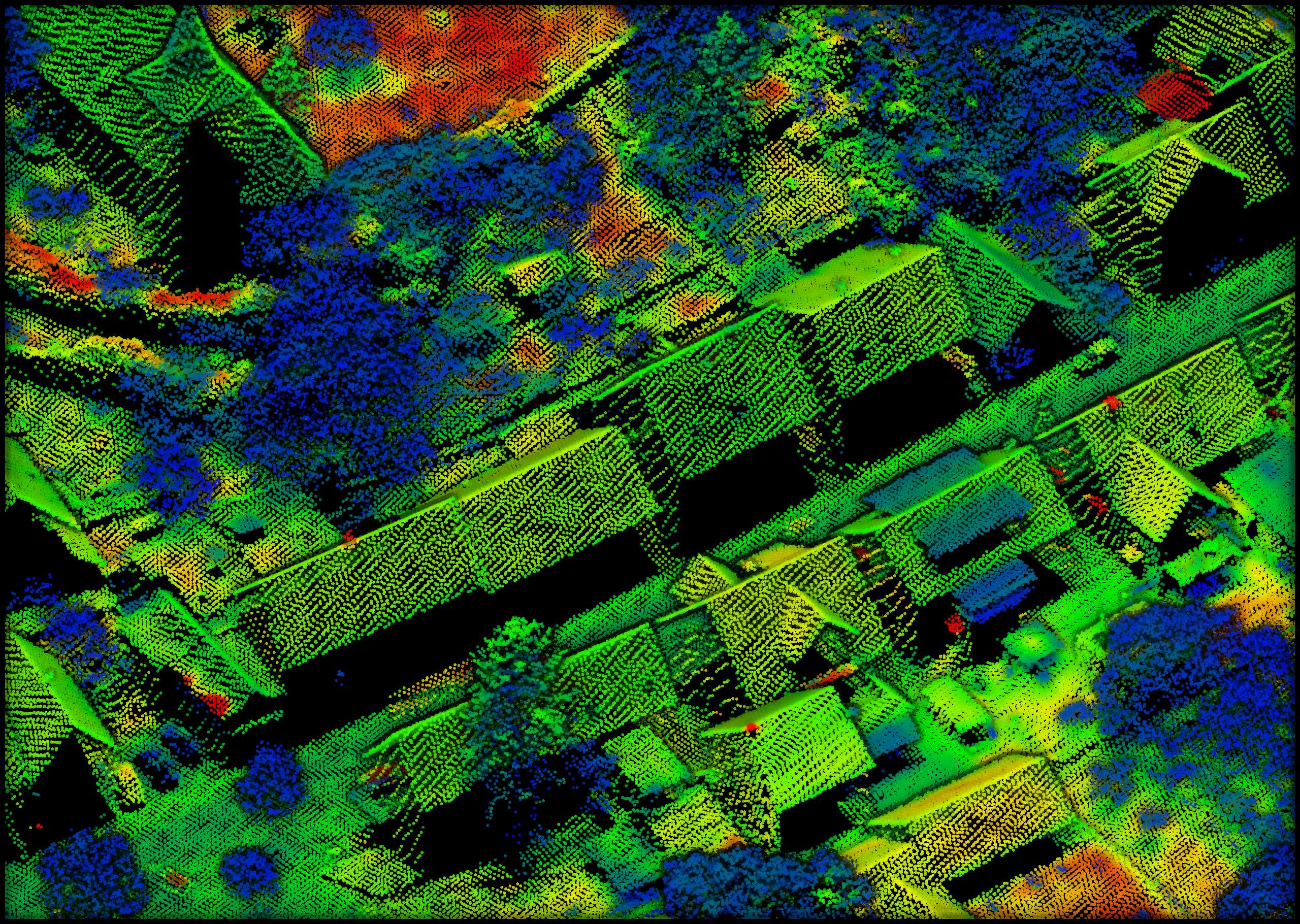}\\
		\vspace{0.2cm}
		\includegraphics[width=0.93\columnwidth,trim=20 20 20 20, clip]{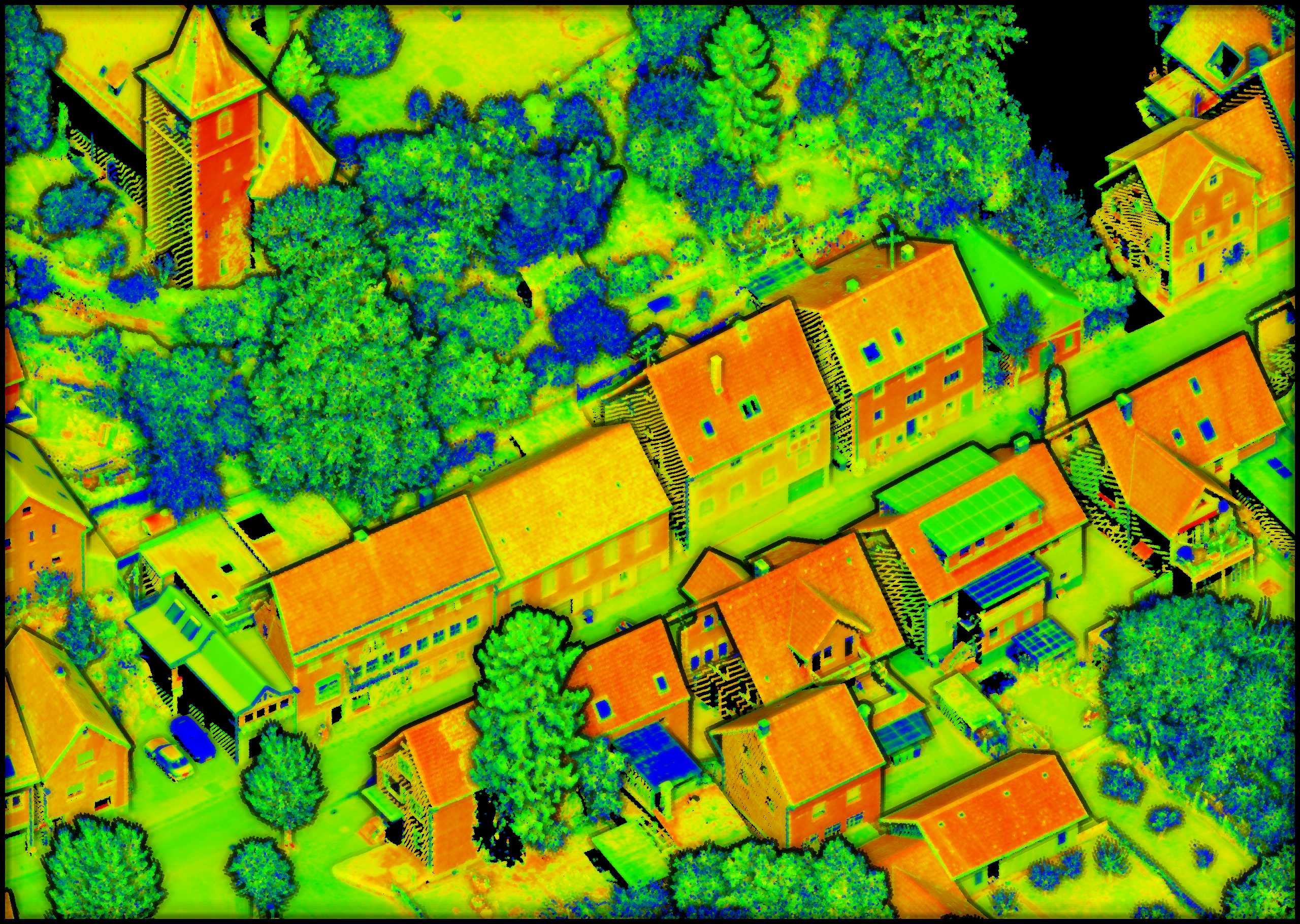}
	\caption{The same subset of the village of Hessigheim captured by a conventional ALS campaign carried out by the the state mapping agency of Baden-Wuerttemberg in 2016 (\textit{top}) and as it is depicted in H3D(PC) (\textit{bottom}).}
    \label{fig:oldvsnew}
\end{center}
\end{figure}

H3D was acquired by a total of $11$ longitudinal (i.e., north-south) strips and several diagonal strips (see Figure~\ref{fig:Density}). Scanner parameters (Pulse Repetition Rate and the mirror's rotation rate) and flying parameters (flying altitude and speed) were set for receiving a point distance of about \unit[5]{cm} both in and across flight direction. Hence, we obtain about \unit[400]{pts/m$^2$} for one single LiDAR strip and about \unit[800]{pts/m$^2$} for the complete point cloud due to strip overlap. As additional strips were flown for further block stabilization, in some areas significantly higher point densities are achieved (see Figure~\ref{fig:Density}). Compared to conventional Airborne Laser Scanning (ALS) flight campaigns applying manned platforms (see Figure~\ref{fig:oldvsnew} \textit{top}), this high-resolution point cloud allows a more comprehensive 3D scene analysis in comparison to existing datasets. For accurate co-registration of acquired strips with respect to available control planes (\cite{Haala2020}), trajectories were corrected by the bias model offered by the \textit{OPALS} software \citep{Pfeifer2014}. In this context, for each strip a constant offset for each trajectory parameter ($\Delta X$, $\Delta Y$, $\Delta Z$, $\Delta \omega$, $\Delta \varphi$, $\Delta \kappa$) was estimated. 

\begin{figure}[htbp]
	\begin{center}
		\subfigure[H3D(PC) - Number of echos]{\includegraphics[width=0.49\columnwidth,trim=10 10 10 10, clip]{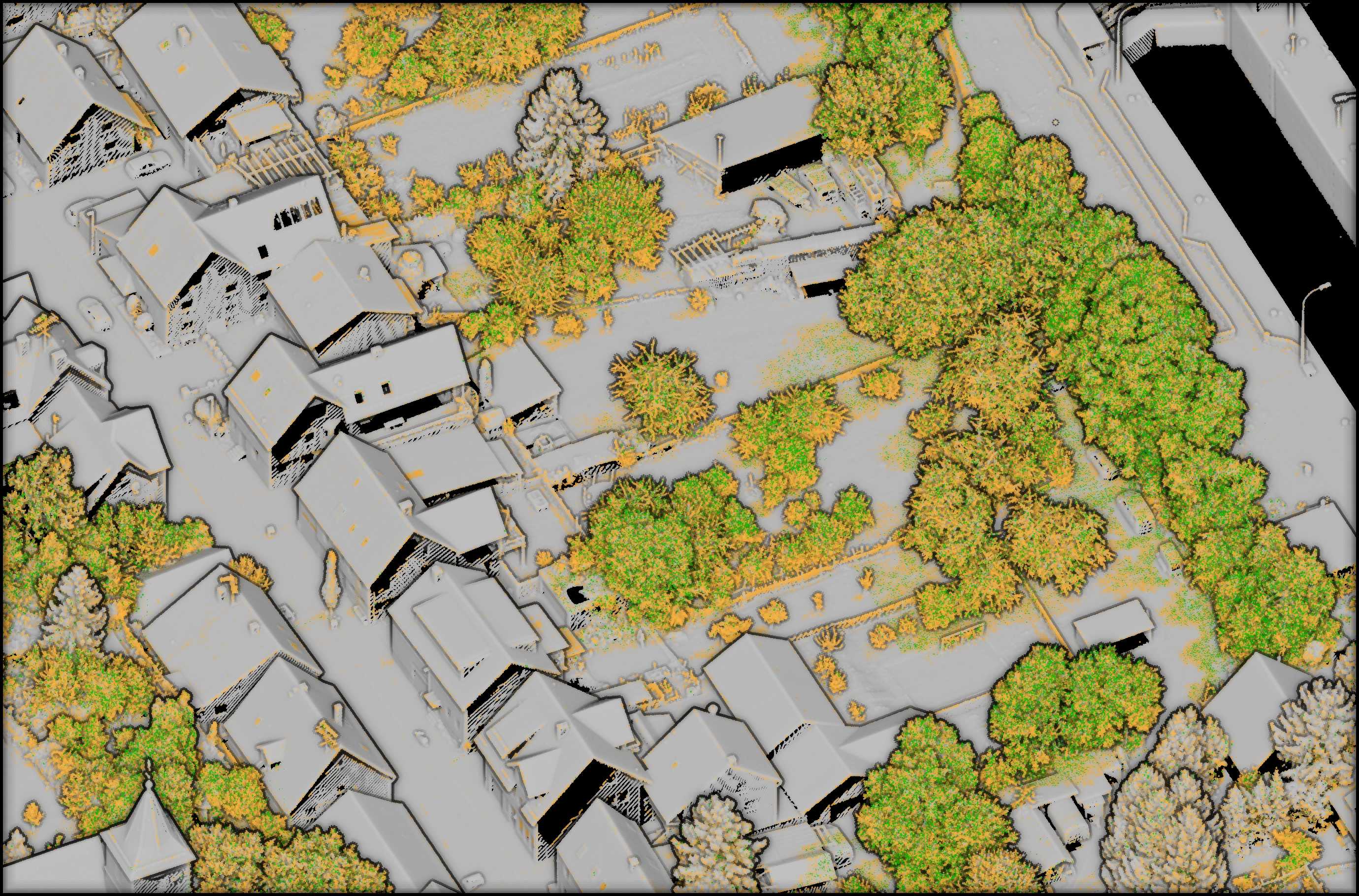}}~ 
		\subfigure[H3D(PC) - Reflectance]{\includegraphics[width=0.49\columnwidth,trim=10 10 10 10, clip]{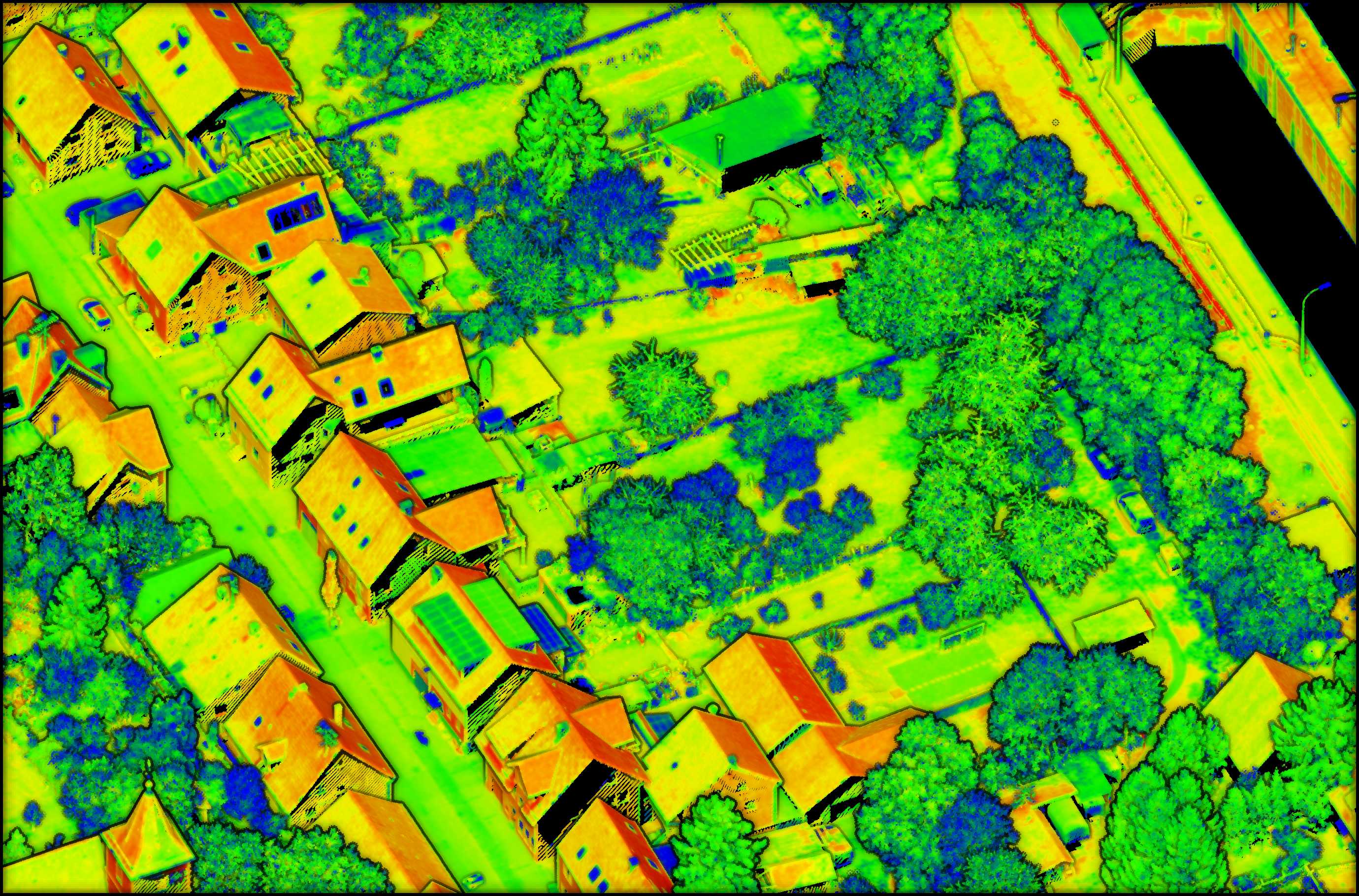}}~ \\
		\subfigure[H3D(PC) - RGB as derived from H3D(Mesh)]{\includegraphics[width=0.49\columnwidth,trim=10 10 10 10, clip]{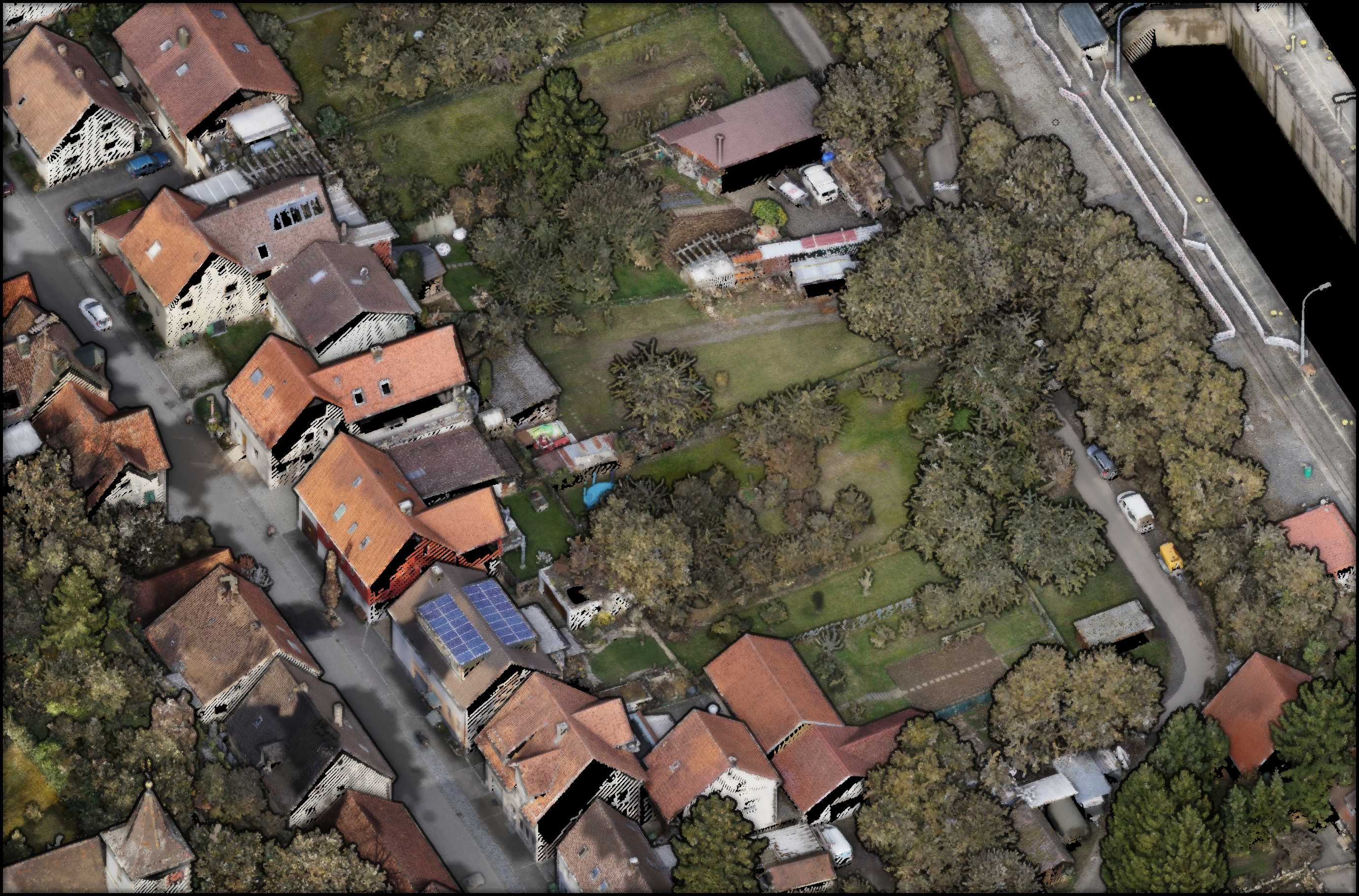}}~ 
		\subfigure[H3D(PC) - Class labels]{\includegraphics[width=0.49\columnwidth,trim=10 10 10 10, clip]{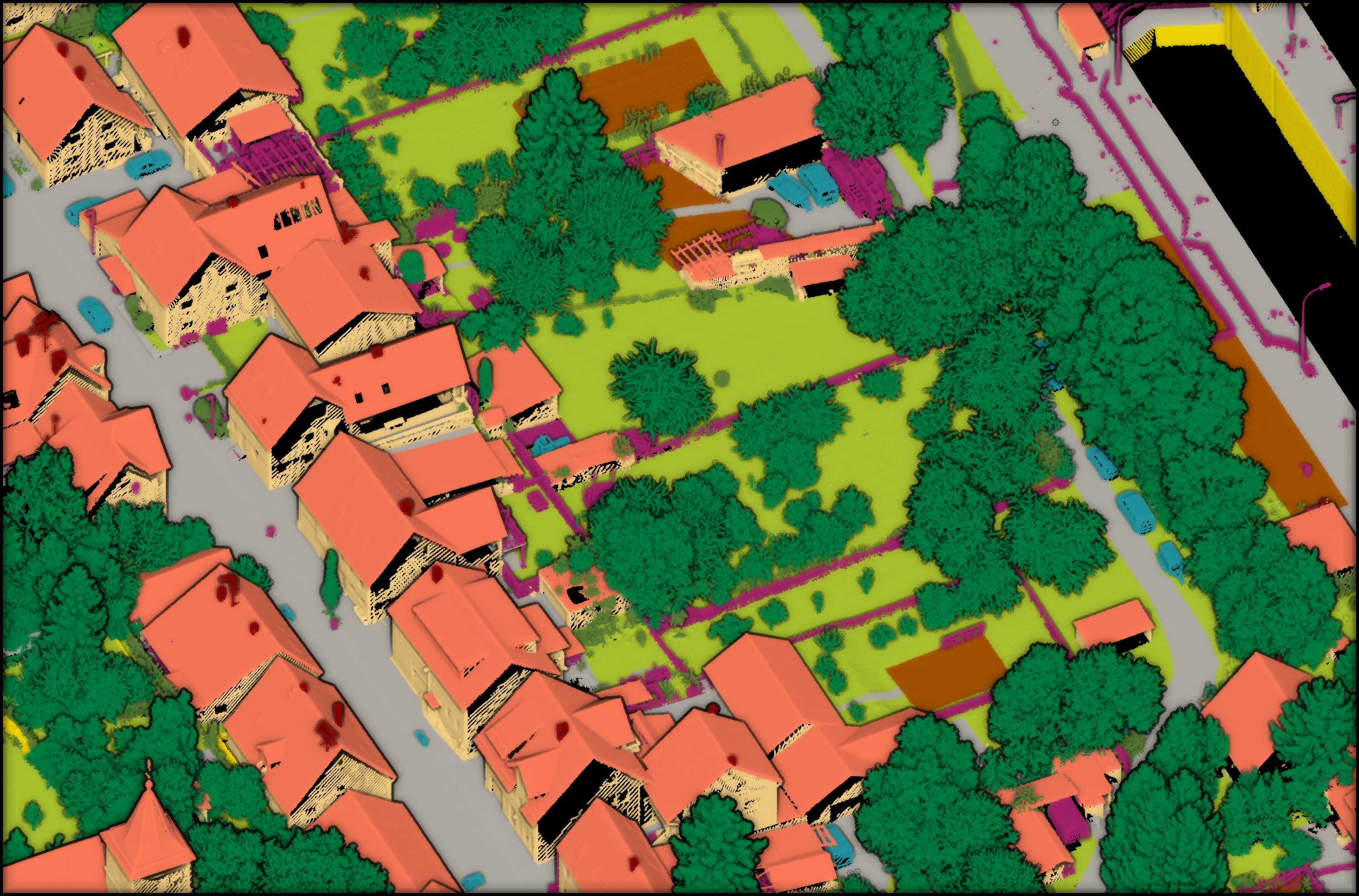}}~ \\
		\subfigure[H3D(Mesh) - RGB]{\includegraphics[width=0.49\columnwidth,trim=10 10 10 10, clip]{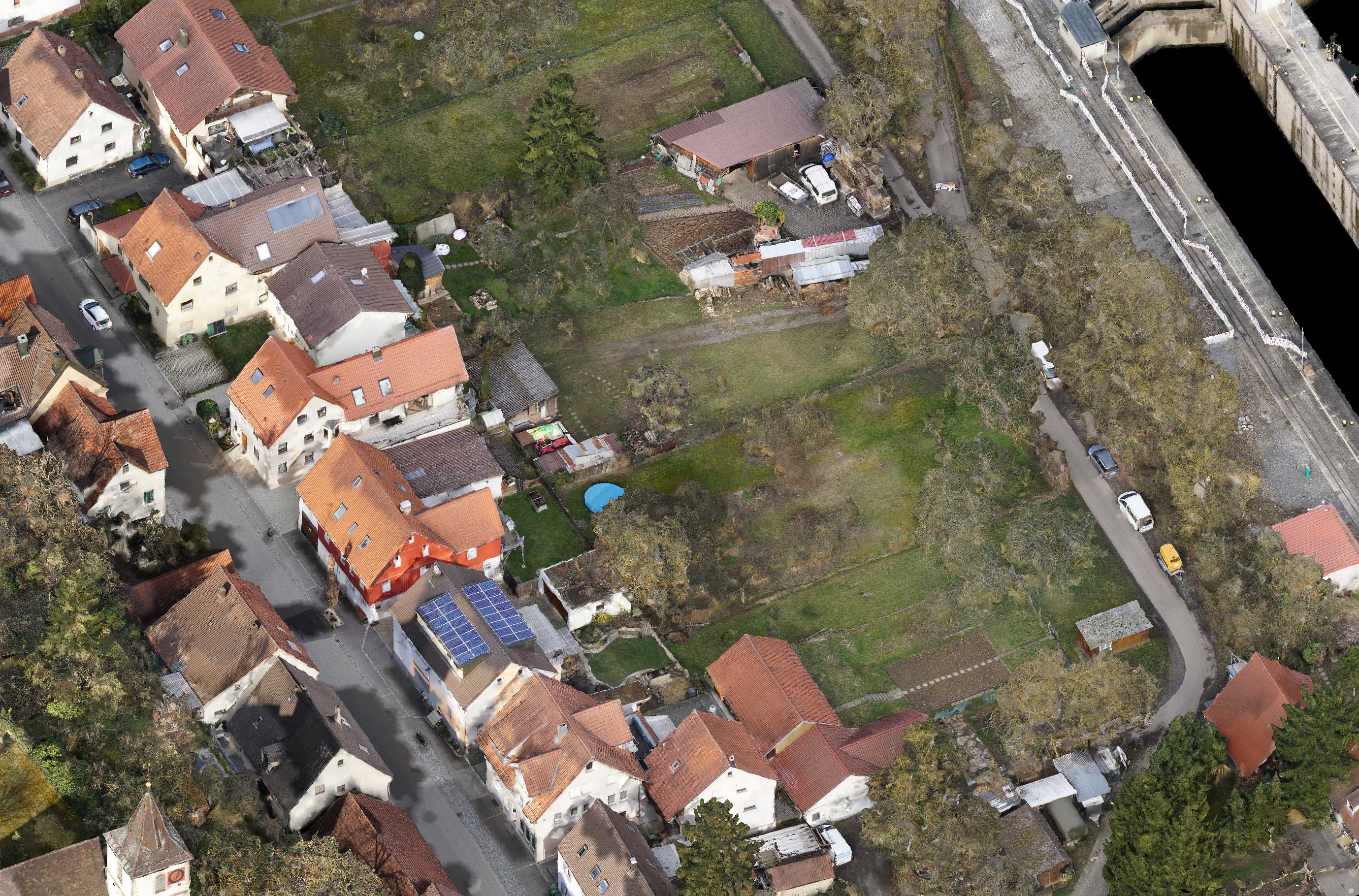}}~ 
		\subfigure[H3D(Mesh) - Class labels transferred from H3D(PC)]{\includegraphics[width=0.49\columnwidth,trim=10 10 10 10, clip]{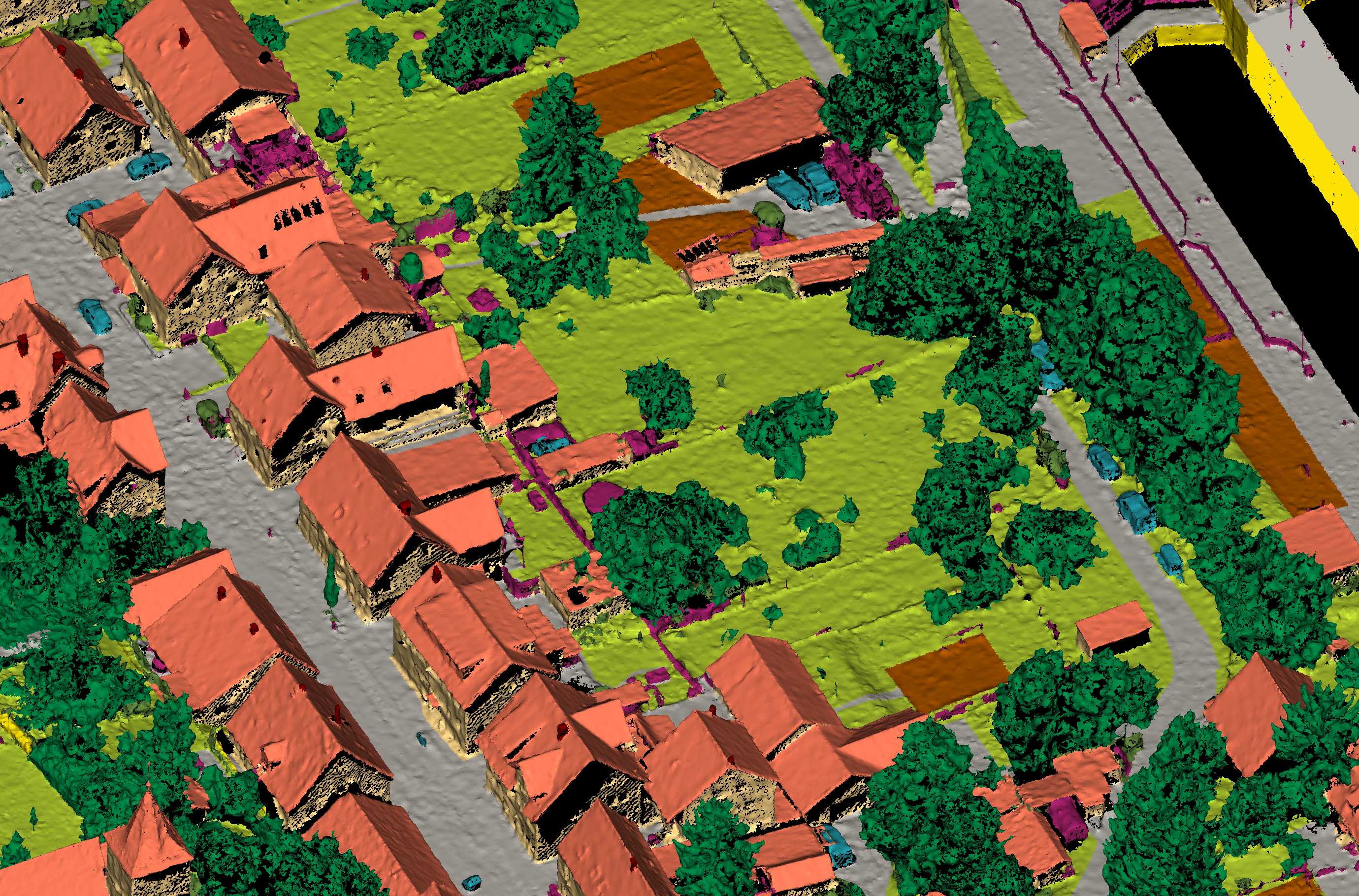}}~ 
		\caption{Available attributes of both modalities of H3D.}
		\label{fig:attributes}
	\end{center}
\end{figure}

Apart from the XYZ coordinates of each point, LiDAR inherent features such as the echo number, number of echos, and reflectance were measured. While up to \unit[6]{echos} were recorded per pulse emitted, the majority of subsequent echos (echo number~$> 1$) are second and third echos (see \textit{yellow} and \textit{green} color in Figure~\ref{fig:attributes} (a)). Instead of the intensity of received echos, we provide reflectance values\footnote[1]{\url{http://www.riegl.com/uploads/tx_pxpriegldownloads/Whitepaper_LASextrabytes_implementation_in-RIEGLSoftware_2017-12-04.pdf}}, which can be interpreted as range corrected intensity. Please note that these values were not corrected for differences in reflectance due to different inclination angles of the laser beam with the illuminated object surface. Reflectance values range from about \unit[-30]{dB} for objects of diffuse reflection properties such as vegetation or asphalt (\textit{dark blue} respectively \textit{light green} points in Figure~\ref{fig:attributes} (b)) and up to about \unit[20]{dB} for objects of directed reflection such as roof or fa\c{c}ade elements (\textit{red} points in Figure~\ref{fig:attributes} (b)).

Point cloud colorization was done in a two-step process. We first derived the mesh as outlined in Section~\ref{sec:Mesh}. Afterwards, we extracted an individual RGB tuple for each LiDAR point from the mesh texture by nearest neighbor transfer. The nearest neighbor interpolation in 3D space is a simple approximation of an occlusion-aware projection of 3D LiDAR points to image space (result is visualized in Figure~\ref{fig:attributes} (c)). 

Additionally, we provide a class label for every point (classes and the annotation will be discussed in Sections~\ref{sec:catalog} and~\ref{sec:GT}). Both plain \textit{ASCII} files and \textit{Las} files are used for data exchange.

\subsection{H3D(Mesh)}\label{sec:Mesh}
We generated the 3D mesh with software \textit{SURE} \citep{Rothermel2012}. For the geometric reconstruction of the scene, both LiDAR data and imagery were used to benefit from their complementary properties \citep{Mandlburger2017}. The fusion of both data sources results in a more complete mesh compared to a mesh derived from images only. For instance, urban canyons are difficult to reconstruct from imagery (due to required visibility in at least two images) but their reconstruction works smoothly for LiDAR data (where one received echo is already sufficient). Furthermore, the oblique images serve for texturing the generated 3D mesh, which allows a realistic representation of vertical faces (e.g., fa\c{c}ades in Figure~\ref{fig:attributes}). 
The mesh data is provided in a tiled manner. 
Each tile is given in both textured and labeled mode. 
For the textured form, each tile consists of i)~an \textit{obj} file describing the geometry and referring to (ii)~the \textit{mtl} file encoding material properties which links to the (iii)~texture atlas that provides textural information (\textit{jpgs}). 
Their labeled counterparts consist of an \textit{obj} file (containing the same geometry as the respective textured version) and a \textit{mtl} which encodes the class properties (i.e., the color-coding).
Therefore, the labeled \textit{obj} files do not require texture atlases since they are pseudo-textured by the class labels. Additionally, we provide Centers of Gravity (CoGs) for each face along with the transferred labels as CoG point cloud. The CoG cloud is available as a plain \textit{ASCII} file enabling simple data handling and data exchange.

\subsection{Class Catalog}\label{sec:catalog}
For H3D(PC) and H3D(Mesh), we employ the same fine-grained class catalog, which is based on V3D but is refined due to H3D's higher point density (and due to the purpose of the Hessigheim project, which is monitoring the shipping lock depicted in Figure~\ref{fig:Splits}). This allows to differentiate more details than in V3D. Hence, we added classes \textit{Urban Furniture}, \textit{Soil/Gravel}, \textit{Vertical Surface} (e.g., found at the shipping lock in Figure~\ref{fig:attributes}) and \textit{Chimney} (see Table~\ref{tab:classes}).
\begin{table}[htpb]
\begin{center}
	\begin{tabular}{cl}
		\toprule
		class ID & class name\\
		\midrule
		0 & {\cellcolor{LowVeg}}\textcolor[rgb]{0,0,0}{Low Vegetation}\\ 
		1 & {\cellcolor{ISurf}}\textcolor[rgb]{0,0,0}{Impervious Surface}\\ 
		2 & {\cellcolor{Vehicle}}\textcolor[rgb]{1,1,1}{Vehicle}\\ 
		3 & {\cellcolor{UFurn}}\textcolor[rgb]{1,1,1}{Urban Furniture}\\ 
		4 & {\cellcolor{Roof}}\textcolor[rgb]{0,0,0}{Roof}\\ 
		5 & {\cellcolor{Facade}}\textcolor[rgb]{0,0,0}{Fa\c{c}ade}\\ 
		6 & {\cellcolor{Shrub}}\textcolor[rgb]{1,1,1}{Shrub}\\ 
		7 & {\cellcolor{Tree}}\textcolor[rgb]{1,1,1}{Tree}\\ 
		8 & {\cellcolor{Soil}}\textcolor[rgb]{1,1,1}{Soil/Gravel}\\ 
		9 & {\cellcolor{VertSurf}}\textcolor[rgb]{0,0,0}{Vertical Surface}\\ 
		10 & {\cellcolor{Chimney}}\textcolor[rgb]{1,1,1}{Chimney}\\ 
		\bottomrule
	\end{tabular}
	\caption{Class catalog of H3D for epoch March 2018.}
	\label{tab:classes}
	\end{center}
\end{table}

\subsection{Generating Ground Truth Data for H3D}\label{sec:GT}
As previously mentioned, the main objective of H3D is to provide labeled multi-temporal and multi-modal datasets for training and evaluation of ML systems for the task of semantic point cloud segmentation. For labeling H3D(PC) (see Section~\ref{sec:PC}), we established a manual process carried out by student assistants, resulting in an annotation as depicted in Figure~\ref{fig:attributes} (d). This classification was generated by extracting point cloud segments with uniform class affiliation (i.e., the point cloud was manually segmented into many small subsets of homogeneous class membership). Segments of each class were afterwards merged to form the semantic segmentation. The whole process was carried out with the help of \textit{CloudCompare} \citep{CloudCompare}. Quality control was accomplished in a two-stage procedure. First, the student assistants checked each other's labels, and finally, the authors verified the results as last instance. Despite the comprehensive quality check, we are aware of the fact that manual annotations are error-prone and label noise cannot be avoided.

For the 3D mesh, we automatically transfer labels from the manually annotated point cloud by a geometry-driven approach that associates the representation entities, i.e., points and faces \citep{Laupheimer2020}. Therefore, the mesh inherits the class catalog (see Table~\ref{tab:classes}) of the point cloud. In comparison to the point cloud representation, the mesh is more efficient because only a small number of faces is required to represent flat surfaces. For this reason, the number of faces is significantly smaller than the number of LiDAR points (see Table~\ref{tab:occurences}). Consequently, several points are commonly linked to the same face. Hence, the per-face label is determined by majority vote of the respective LiDAR points. However, due to structural discrepancies, some faces remain unlabeled because no points can be associated with them (e.g., absence of LiDAR points or geometric reconstruction errors). These faces are marked by the pseudo class label~$-1$. 
Unlabeled faces cover about \unit[40]{\%} of the entire mesh surface. 
As can be seen from Figure~\ref{fig:Splits}, the majority of the unlabeled area (\unit[99.7]{\%}) belongs to parts where the mesh exceeds the labeled point cloud (due to the tiled mesh structure). 
For the overlap of LiDAR and mesh (i.e., the relevant data), \unit[84]{\%} of the surface carries an annotation.


\begin{sidewaystable}
\begin{adjustbox}{scale=0.76,center}
    \begin{tabular}{llllccccccccccc}
        \toprule
        & & & &\multicolumn{11}{c}{Classes}\\ \cmidrule{5-15}
        Modality & Split & total & & \textit{Low Veg.} & \textit{I. Surf} & \textit{Vehicle} & \textit{U. Furn.} & \textit{Roof} & \textit{Fa\c{c}ade} & \textit{Shrub} & \textit{Tree} & \textit{Soil} & \textit{V. Surf.} & \textit{Chimney}\\
        \midrule
        \multirow{4}{*}{H3D(PC)} 
            & \multirow{2}{*}{Train}         
                & \multirow{2}{*}{\unit[59445106]{Pts}}   
                        & abs.      & 21375614   & 10419635    & 258032    & 1159205  & 6279431    & 1198227   & 1077141   & 8086818   & 8590706 & 974976 & 25321\\
                    & & & rel. [\%] & 35.96   & 17.53    & 0.43    & 1.95  & 10.56    & 2.02   & 1.81   & 13.60   & 14.45 & 1.64 & 0.04\\
            & \multirow{2}{*}{Validation} 
                & \multirow{2}{*}{\unit[14464248]{Pts}}    
                        & abs.      & 3738743   & 3212988    & 183263    & 455389  & 3052150    & 551996   & 341579   & 2218551   & 592510 & 101243 & 15836\\
                    & & & rel. [\%] & 25.85   & 22.21    & 1.27    & 3.15  & 21.10    & 3.82   & 2.36   & 15.34   & 4.10 & 0.70 & 0.11\\
        \midrule                                                    
        \multirow{8}{*}{H3D(Mesh)} 
            & \multirow{4}{*}{Train} 
                & \multirow{2}{*}{\unit[923663]{Faces}} 
                        & abs. & 1333755 & 692922 & 44803 & 220790 & 455718 & 258606 & 184263 & 1258053 & 498072 & 2143730 & 5326 \\
                    & & & rel. [\%]& 25.81 & 13.41 &  0.87 & 4.27 & 8.82 & 5.01 &  3.57 & 24.35 &  9.64 &  4.15 & 0.10 \\
              & & \multirow{2}{*}{\unit[136836.45]{m$^2$}}  
                        & abs.     & 22146.7 & 12502.6 &    660.0 &    2990.7 &    7346.3 &    4123.1 &    2186.9 &    16935.5 &    9754.2 &    3797.2 &    59.4 \\
                    & & & rel. [\%]& 26.84 & 15.15 &  0.80 &  3.63 &  8.90 & 5.0 &  2.65 & 20.53 & 11.82 &  4.60 & 0.07 \\
            & \multirow{4}{*}{Validation} 
                & \multirow{2}{*}{\unit[2577554]{Faces}}    
                        & abs.     & 266940 & 236243 & 28836 & 90050 & 245665 & 135410 & 59385 & 322881 & 39142 & 25936 & 3812\\
                    & & & rel. [\%]& 18.36 & 16.24 & 1.98 & 6.19 & 16.89 & 9.31 &  4.08 & 22.20 &  2.69 &  1.78 & 0.26\\
              & & \multirow{2}{*}{\unit[37928.05]{m$^2$}}   
                        & abs.     & 4443.6 &    4203.6 &    439.9 &    1276.1 &    4016.1 &    2223.7 &    829.8 &    4694.4 &    674.4 &    472.5 &    41.1 \\
                    & & & rel. [\%]& 19.06 & 18.03 &  1.89 &  5.47 & 17.23 & 9.54 &  3.60 & 20.13 &  2.89 &  2.03 & 0.18 \\
        \bottomrule
    \end{tabular}
\end{adjustbox}
    \caption{Class frequencies in the training and validation sets both for H3D(PC) and H3D(Mesh). For H3D(Mesh) we further present the class distribution in terms of covered area. Unlabeled faces are not considered for these statistics.}
    \label{tab:occurences}
\end{sidewaystable}

\subsection{Data Splits}\label{sec:splits}
\begin{figure}[htbp]
	\begin{center}
		\includegraphics[width=0.86\columnwidth,trim=165 220 180 0, clip]{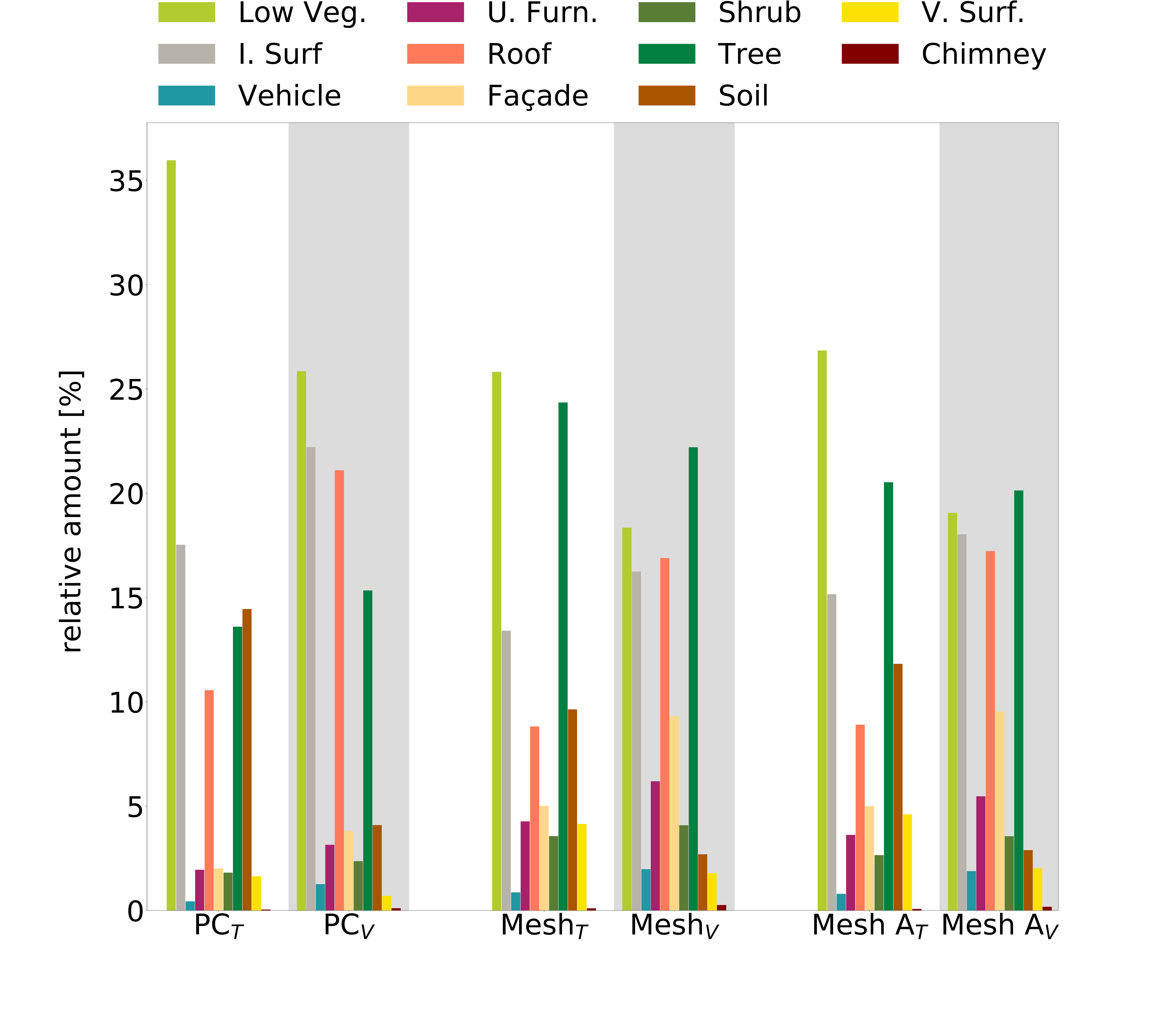}
		\caption{Class distributions of the training set $T$ and the validation set $V$ both for H3D(PC) and H3D(Mesh). For the mesh, the relative class distribution is given in terms of i) the face count and ii) the area $A$ of faces. Unlabeled faces are not considered in this figure.}
		\label{fig:class_distribution}
	\end{center}
\end{figure}

The datasets of all epochs are split into a distinct training, validation, and test area (see Figure~\ref{fig:Splits}). The splits are identical in both modalities and chosen in accordance with the mesh tiling. Since H3D is designed as a benchmark, labels of the test set are not disclosed to the participants of the benchmark. Whereas labels of the training and validation sets can be used by participants as desired, we recommend utilizing the pre-defined splits. Detailed statistics of class frequencies in the training and validation sets can be found in Table~\ref{tab:occurences} for both modalities (points vs. faces/CoGs) and are visualized in Figure~\ref{fig:class_distribution}. For the mesh, we additionally provide the area each class covers, measured by the area of faces assigned to the corresponding class.

In case of the point cloud, the relative number of points for classes covering large areas (such as \textit{Low Vegetation}, \textit{Impervious Surface}, \textit{Tree} and \textit{Roof}) is naturally the highest. Regarding class \textit{Tree}, the number of points is further increased since multiple echos were received for one laser beam. The most underrepresented classes are \textit{Vehicle} and \textit{Chimney}. Although the relative class frequencies for the mesh are similar, we can observe that there are fewer instances of classes corresponding to the ground  (\textit{Low Vegetation}, \textit{Impervious Surface} and \textit{Soil/Gravel}), which is due to large face elements used for representing such planar surfaces.  On the other hand, the relative number of instances of class \textit{Tree} is increased, because vegetation surfaces are characterized by a high roughness and, thus  are typically approximated by a large number of surface elements (i.e., faces.)

\subsection{Benchmark Challenge}\label{sec:task}
In contrast to V3D, the H3D benchmark challenge is twofold in terms of data representation. For both H3D(PC) and H3D(Mesh), we offer participants to use the training and validation dataset for developing ML approaches for supervised classification and then to apply those to the test dataset. Predicted labels are to be returned to the authors for evaluation (see Section~\ref{sec:metrics}). For the mesh, the predictions are to be returned to the authors as labeled CoG cloud (the respective plain \textit{ASCII} file of CoGs is provided by the authors). The authors will match the predicted per-face labels with the corresponding faces of the \textit{obj} files. Simply put, the CoG cloud is utilized as an efficient link to the underlying mesh structure to keep the memory footprint of submitted data low. The evaluation itself will be done on the mesh (\textit{obj} files).

\subsection{Evaluation Metrics}\label{sec:metrics}
Semantic segmentation results submitted by participants of the benchmark are evaluated by the H3D team by means of the derived confusion matrices. In order to obtain performance metrics for individual classes, the number of True Positives~($TP$), the number of False Positives~($FP$), and the number of False Negatives~($FN$) are determined for each class~$c$. These numbers are used for determining Precision~($P$) and Recall~($R$) (Equation~\ref{equ:1}~\&~\ref{equ:2}). Additionally, we derive a F1-score as the harmonic mean of $P$ and $R$ for each class (Equation~\ref{equ:3}) \citep{Goutte2005}.
\begin{equation}\label{equ:1}
  \begin{aligned}
		P_c = \frac{TP_c}{TP_c+FP_c}
  \end{aligned}
\end{equation}
\begin{equation}\label{equ:2}
  \begin{aligned}
		R_c = \frac{TP_c}{TP_c+FN_c}
  \end{aligned}
\end{equation}
\begin{equation}\label{equ:3}
  \begin{aligned}
		\text{F}1_c = \frac{2 \cdot P_c \cdot R_c}{P_c+R_c}
  \end{aligned}
\end{equation}

To describe the total performance of a classifier, we combine individual class scores by computing i) the Overall Accuracy (\mbox{$OA=\sum_c TP_c/N$}, with $N$ being the total number of labeled instances) and ii) the mean F1-score (macro-F1). These measures are determined both for H3D(PC) and H3D(Mesh). In case of the latter, the evaluation is based on the covered area of correctly / incorrectly classified faces (as provided as CoG cloud by the participants).

\section{Baselines}\label{sec:baselines}
We initialize the H3D benchmark challenge by providing two baseline solutions for semantic segmentation of H3D. On one hand, we apply a conventional Random Forest (RF) classifier \citep{Breiman2001} relying on hand-crafted features (see Section~\ref{sec:RF}) and on the other hand, we use a Sparse Convolutional Network (SCN) as end-to-end learning approach (see Section~\ref{sec:SCN}).

\subsection{Random Forest}\label{sec:RF}
For semantic segmentation of the point cloud, we compute geometric features as proposed in \cite{Weinmann2015} and in \cite{Chehata2009} by estimating the structural tensor for a local neighborhood of each point. After extracting the Eigenvalues of that tensor, we can determine the characteristics of the respective point distribution by forming different ratios of Eigenvalues \citep{Weinmann2015}. As computing Eigenvalues and Eigenvectors eventually means to fit a plane to the local point set, we can further enhance our feature vector by taking into account the orientation of these planes. Furthermore, we consider height-based features by determining the height above ground (i.e., above the Digital Terrain Model (DTM) level; DTM is derived by \textit{SCOP++} \citep{Pfeifer2001}) for each LiDAR point. In addition to purely geometric features, LiDAR-inherent features such as echo ratio and intensity of the received echo are also used for the semantic segmentation. In order to establish a multi-scale approach and to analyze features on different levels of abstraction, we follow the recommendation of \cite{Weinmann2018a} and compute each feature for spherical neighborhoods of radii $r=1,2,3$ and \unit[5]{m}. To account for the high density of H3D, we expand the feature vector of each point by additionally deriving all geometric features for radii $r=0.125$, $0.25$, $0.5$ and \unit[0.75]{m}.

To obtain mesh features, we follow the approach of \cite{Tutzauer2019} and encode each face by its CoG. 
In this way, we can on one hand compute all aforementioned geometric features for our CoG cloud. On the other hand, we preserve features of the mesh geometry by assigning mesh-inherent features such as face area, face density, and normal orientation to the respective faces. Furthermore, we transfer LiDAR-specific features to the mesh representation by the approach presented in \cite{Laupheimer2020}. 

For both the point cloud and the mesh, we additionally incorporate radiometric features. For this purpose, RGB tuples are converted to HSV color space and used together with Gaussian smoothed color values for the aforementioned spatial neighborhoods. Since a multitude of HSV tuples is encoded in each face, we additionally calculate the HSV variance for each face. 

Based on these features, a RF model is trained for H3D(PC) and H3D(Mesh). Prediction results for the test set can be found in Table~\ref{tab:baseline} (see discussion of results in Section~\ref{sec:discussion}). The RF models are parametrized by $100$ binary decision trees with a maximum depth of $18$. \cite{Niemeyer2014} have shown that pure pointwise results of the RF classifier can be further refined by a Markov Random Field (MRF). Therefore, we enhance our RF by an a posteriori probability-aware MRF-like smoothing (a posteriori probability is used as unary potential; points within \unit[0.5]{m} radius are considered for the regularization term). 
\begin{table*}[htbp]
\begin{adjustbox}{max width=\textwidth}
	\begin{tabular}{llccccccccccccc}
		\toprule
		& &\multicolumn{11}{c}{F1-scores [\%]}\\ \cmidrule{3-13}
		Modality & Method & \textit{Low Veg.} & \textit{I. Surf} & \textit{Vehicle} & \textit{U. Furn.} & \textit{Roof} & \textit{Fa\c{c}ade} & \textit{Shrub} & \textit{Tree} & \textit{Soil} & \textit{V. Surf.} & \textit{Chimney} & mF1 [\%] & OA [\%]\\
		\midrule
		\multirow{2}{*}{H3D(PC)} & RF   &	90.36 & 88.55 & 66.89 & 51.55 & 96.06 & 78.47 & 67.25 & 95.91 & 47.91 & 59.73 & 80.65 & 74.85 & 87.43 \\
								 & SCN &	92.31 & 88.14 & 63.51 & 57.17 & 96.86 & 83.19 & 68.59 & 96.98 & 44.81 & 78.20 & 73.61 & 76.67 & 88.42 \\
		\midrule
		\multirow{2}{*}{H3D(Mesh)}  & RF   &	89.35 & 89.06 & 61.38 & 57.65 & 93.29 & 82.16& 69.27 & 96.09 & 48.85 & 62.58 & 76.43 & 75.10 & 86.53 \\
								    & SCN & 89.82 & 83.66 & 61.05 & 52.24 & 87.09 & 81.01 & 59.75 & 95.06 & 51.00 & 56.61 & 70.22 & 71.59 & 83.73 \\
		\bottomrule
	\end{tabular}
	\end{adjustbox}
	\caption{Baseline results of semantic segmentation for both H3D(PC) and H3D(Mesh). 
	For the mesh, we report the performance metrics weighted by the covered surface.}
	\label{tab:baseline}
\end{table*}

\subsection{Sparse Convolutional Network}\label{sec:SCN}
As deep learning has become a de-facto standard in most fields of pattern recognition, we also include a neural network in our baselines. In particular, we employ a 3D CNN in U-Net form with submanifold sparse convolutional layers \citep{Graham2018} to account for the typical spatial distribution of ALS point clouds. The network's general layout and training regime is described in \cite{SchmohlSoergel2019a}. It consists of $3$ downsampling levels and $21$ convolutional layers in total. 
The input data (point cloud or mesh) is discretized to sparse 3D voxel grids of \unit[25]{cm} side length and contains just the raw attributes (i.e., no features are computed as for the RF). 
In case of the point cloud, these are the measured point attributes (i.e., echo number, number of echos, reflectance \& RGB values) as described in Section~\ref{sec:PC}. 
For the mesh, voxels are derived from the CoG point cloud and attributed with the texture information (RGB) only.
We tested also a configuration with additional normal information, since this is the most basic native mesh feature besides RGB. 
However, the achieved performance is slightly worse ($-0.6$ and $-2.34$ percent points for OA and mF1-score respectively).
For evaluation, the inferred voxel labels are transferred back to the enclosed points or faces. Results are also reported in Table~\ref{tab:baseline}.

\subsection{Discussion of Baseline Results}\label{sec:discussion}
Within this chapter, we analyze the performance of our two baseline classifiers (see~Section\ref{sec:discussionPC} and~Section{sec:discussionMesh} respectively) for both H3D(PC) and H3D(Mesh) in order to develop a better understanding of challenges of H3D and to kick-off the benchmark competition. 

\subsubsection{H3D(PC)}\label{sec:discussionPC}
\begin{figure}[htbp]
	\begin{center}
		\subfigure[Labels predicted by RF]{\includegraphics[width=0.49\columnwidth,trim=10 10 10 10, clip]{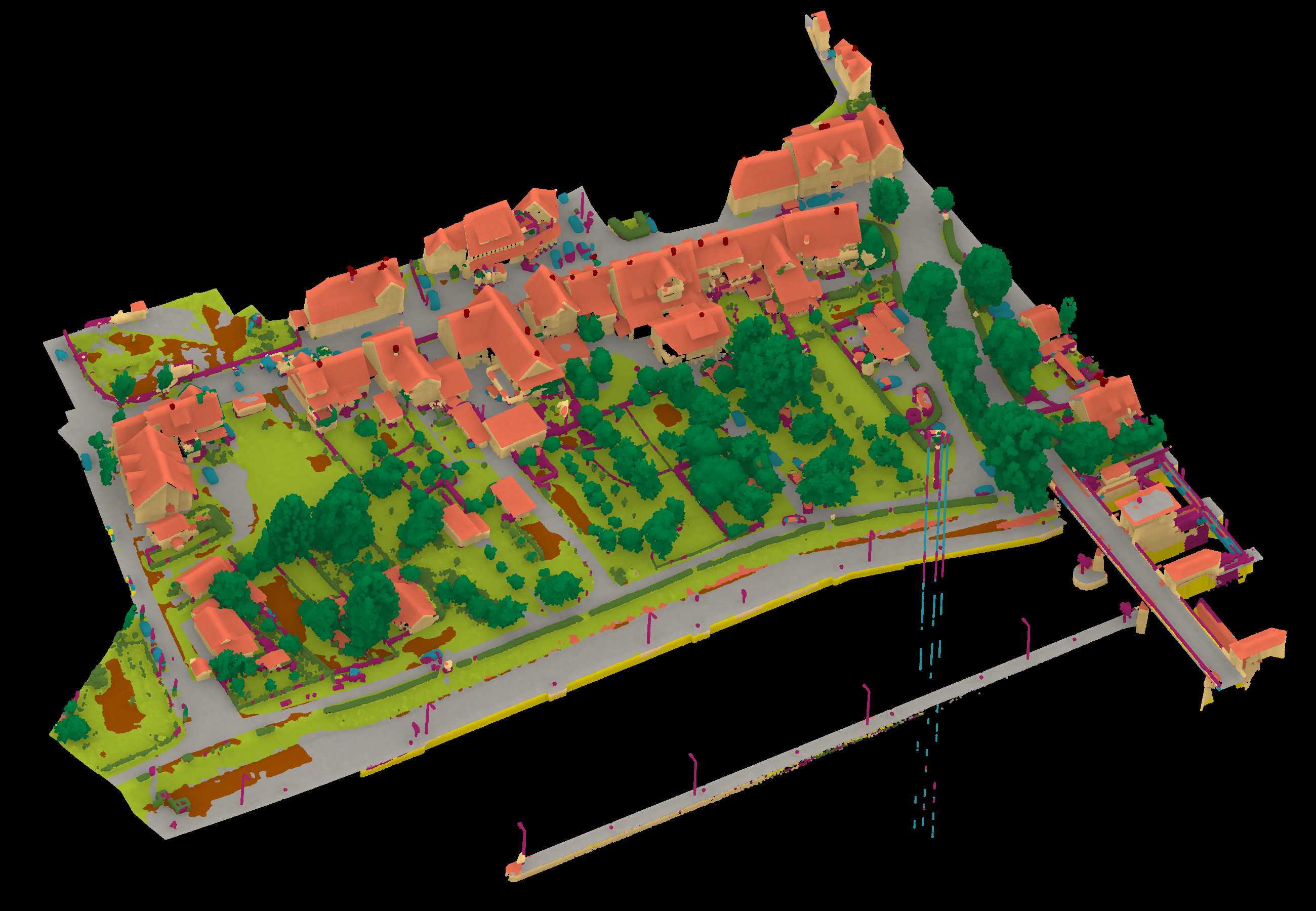}}~ 
		\subfigure[Labels predicted by SCN]{\includegraphics[width=0.49\columnwidth,trim=10 10 10 10, clip]{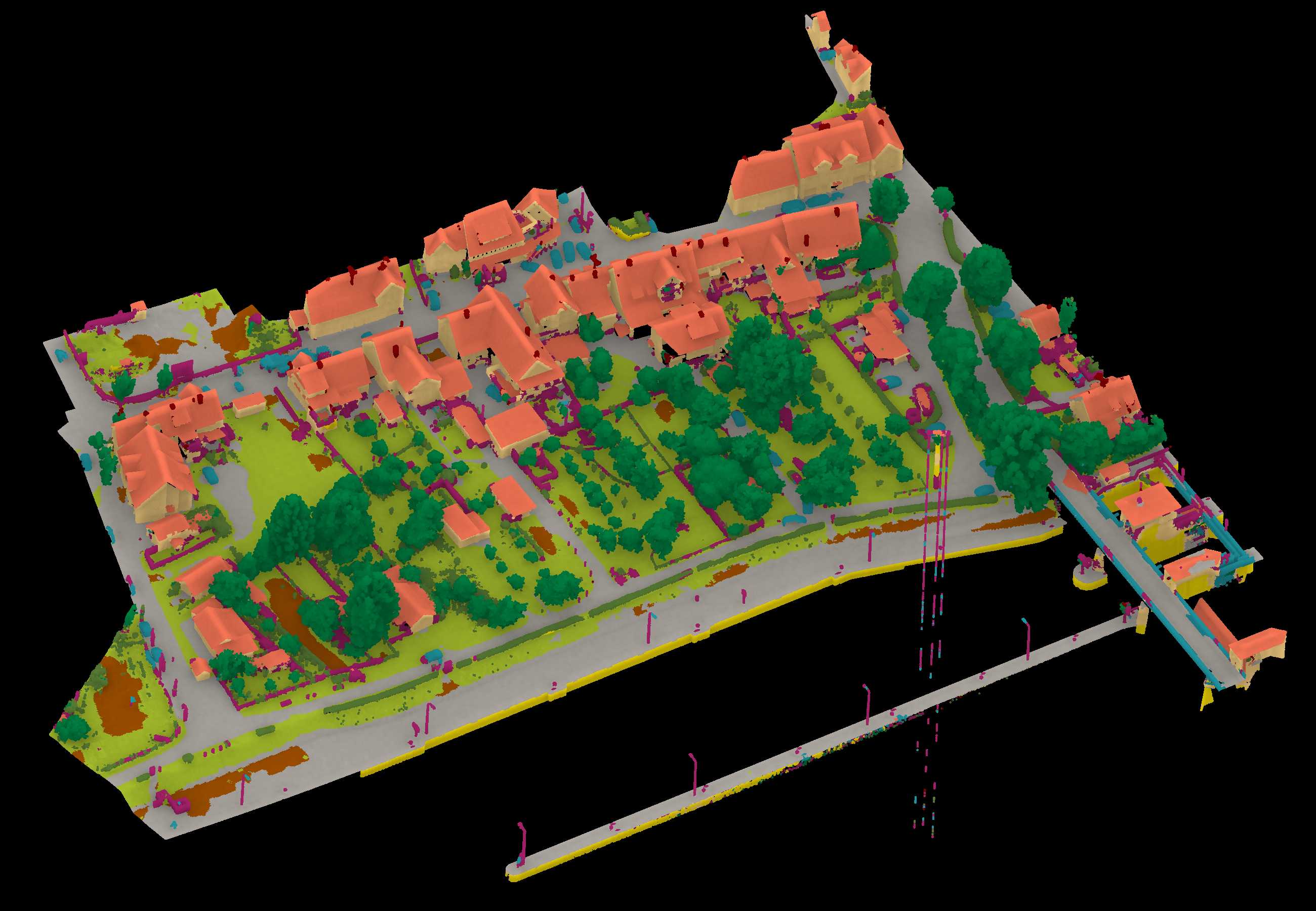}}~ \\
		\subfigure[Labels predicted by RF - close up]{\includegraphics[width=0.49\columnwidth,trim=10 10 10 10, clip]{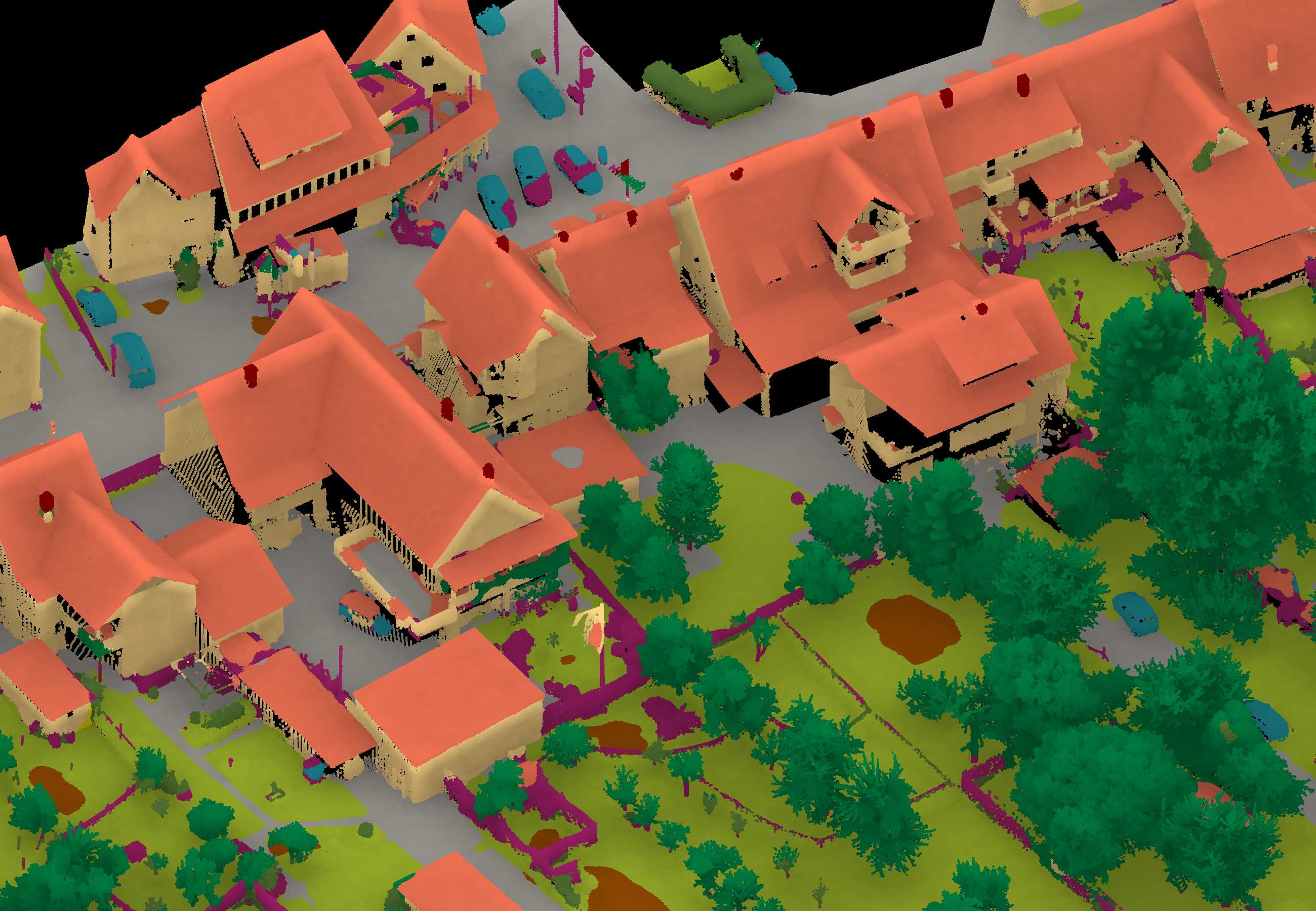}}~ 
		\subfigure[Labels predicted by SCN - close up]{\includegraphics[width=0.49\columnwidth,trim=10 10 10 10, clip]{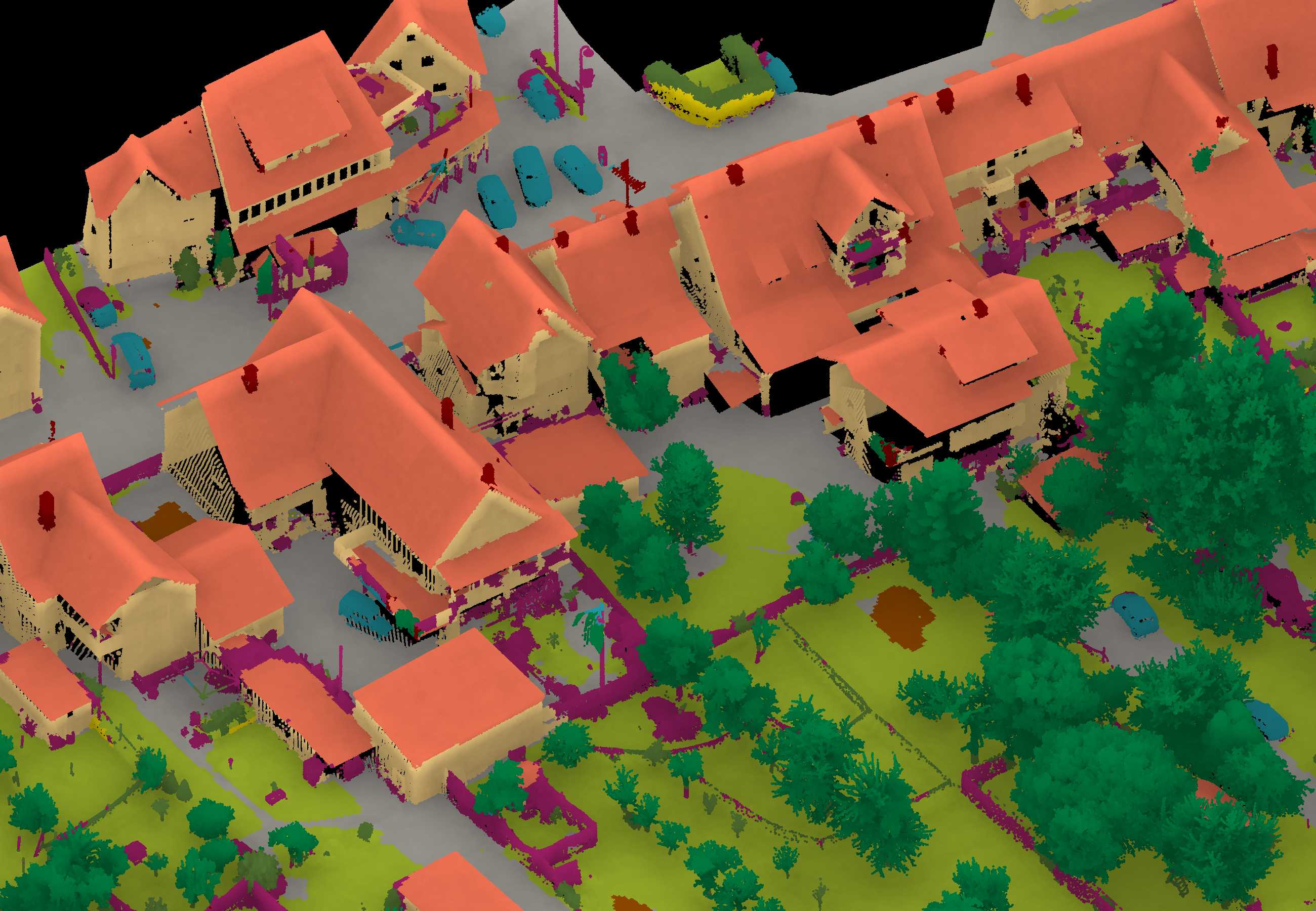}}~ \\
		\subfigure[Normalized confusion matrix for the RF classifier]{\includegraphics[width=0.49\columnwidth,trim=0 0 0 22, clip]{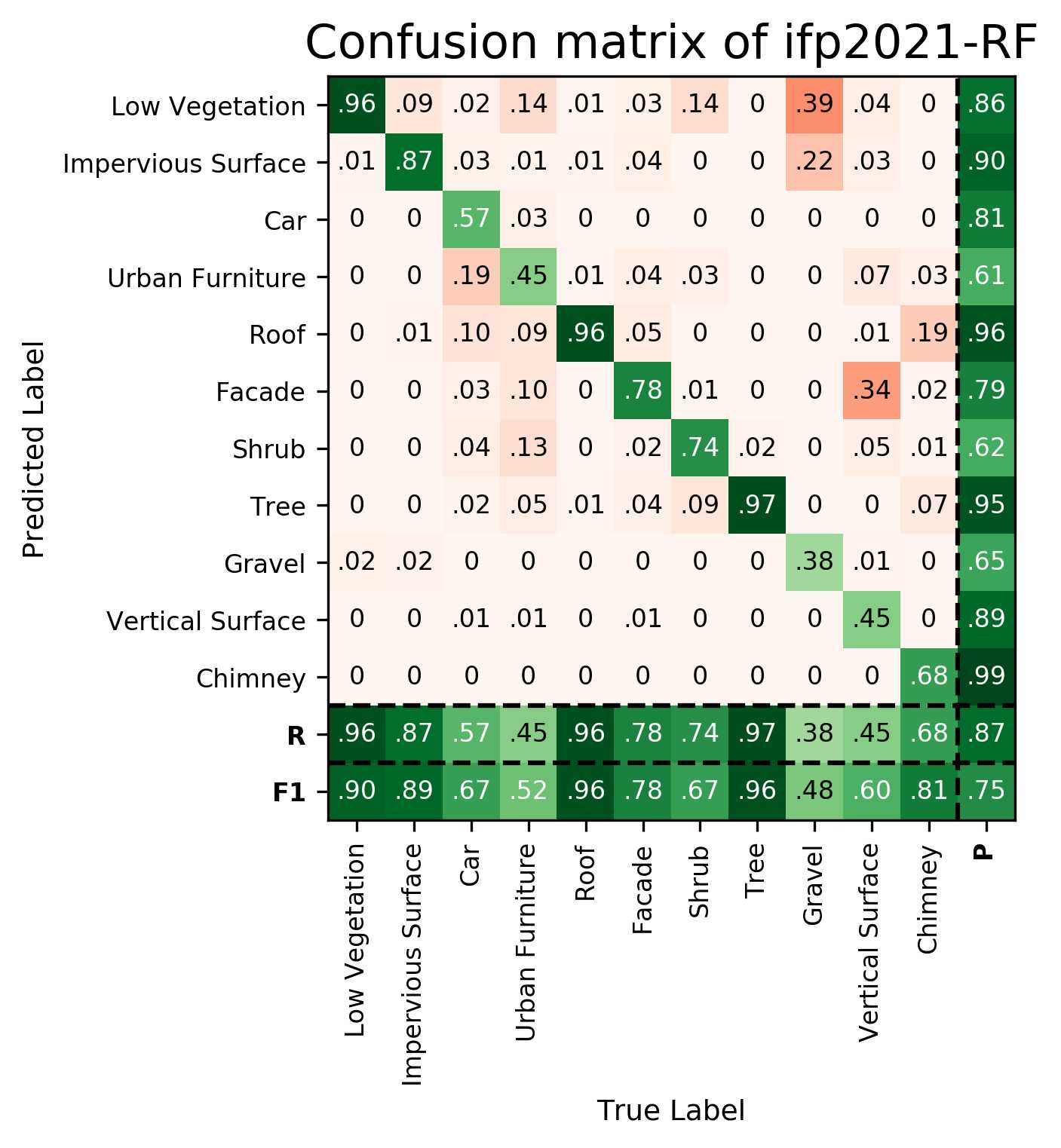}}~ 
		\subfigure[Normalized confusion matrix for the SCN classifier]{\includegraphics[width=0.49\columnwidth,trim=0 0 10 22, clip]{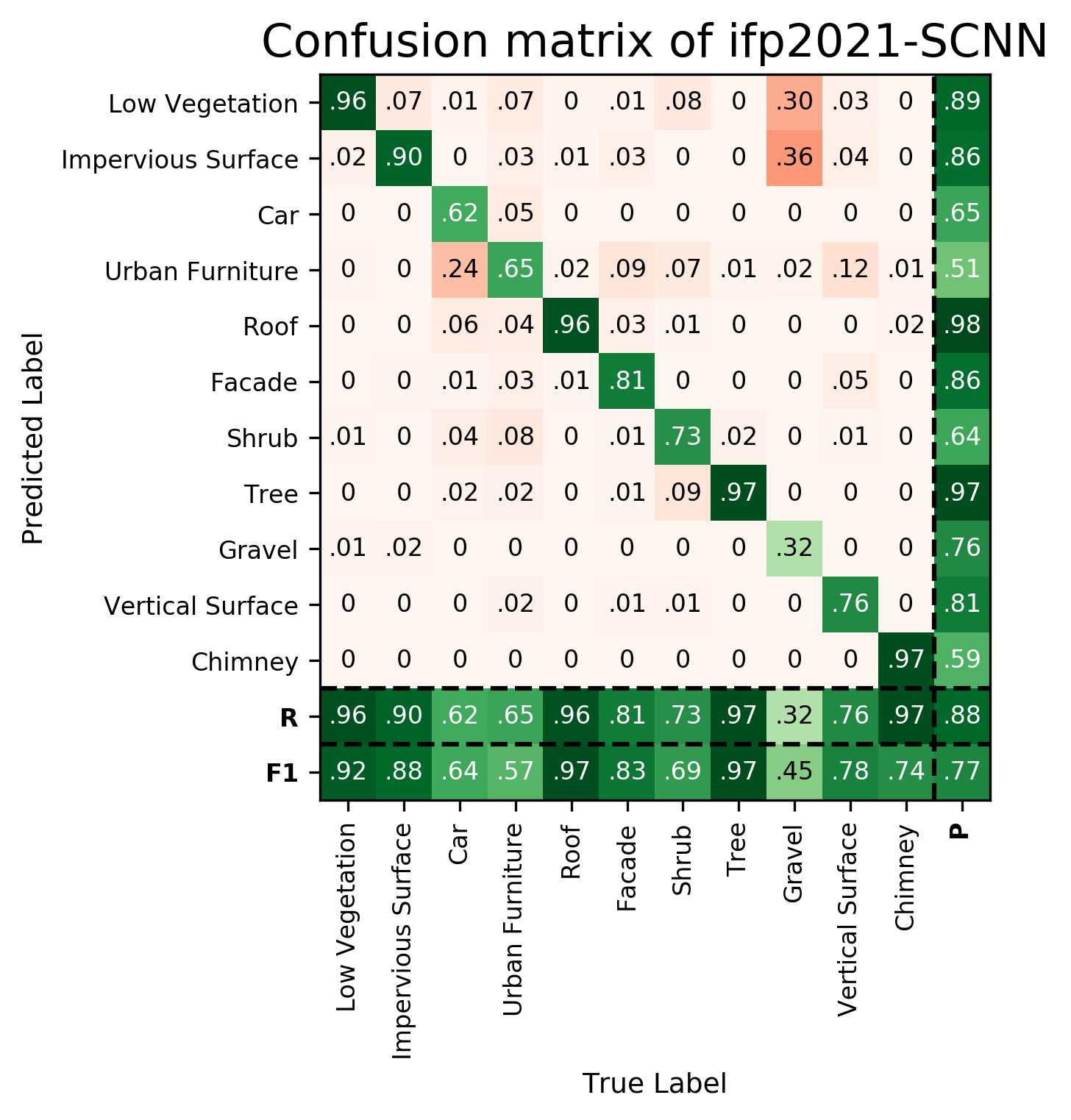}}~ \\
		\caption{Results of semantic segmentation on the test set of H3D(PC) for both our RF (\textit{left}) and our SCN classifier (\textit{right}).}
		\label{fig:predPC}
	\end{center}
\end{figure}
Generally, we can observe from Table~\ref{tab:baseline} that the RF and the SCN classifiers perform similarly well, with the SCN slightly better in terms of the OA and the mean F1-score. We want to stress that reached accuracies already exceed the best result of the V3D challenge (OA of \unit[85.2]{\%} by \cite{Zhao2018}), which underlines the superiority of high-resolution data. Both baseline solutions (RF \& SCN) achieve similar results for the ground classes \textit{Low Vegetation} and \textit{Impervious Surface}. However, performance for class \textit{Soil/Gravel} is rather poor in both cases for there is often confusion with other ground classes (see Figure~\ref{fig:predPC}). Since points of all ground classes incorporate similar geometric properties (e.g., similar normals and smooth surfaces), distinguishing these classes is only possible with the help of radiometric features such as reflectance or color information (see Figure~\ref{fig:attributes}). Whereas this performs well for \textit{Low Vegetation} vs. \textit{Impervious Surface}, segmenting points of \textit{Soil/Gravel} is rather demanding due to similar radiometric properties to \textit{Low Vegetation} (similar to bare soil) and \textit{Impervious Surface} (similar to debris and gravel).

Similar radiometric and geometric properties are also the reason for the confusion of \textit{Shrub} with \textit{Tree} (greenish color in both cases and rough surfaces). Nevertheless, the extraction of tree points succeeds quite well, probably due to the distinctive multi-echo ability of the employed sensor (see Figure~\ref{fig:attributes} (a)). Regarding buildings, roofs can be extracted successfully, but the detection of fa\c{c}ades seems to be more demanding for both classifiers (see Table~\ref{tab:baseline}). This may be caused by the presence of fa\c{c}ade furniture (such as balconies with handrails and outdoor furniture differing from smooth fa\c{c}ades), which is similar to urban furniture. The confusion matrices in Figure~\ref{fig:predPC} further indicate that points of class \textit{Urban Furniture} are spread across many other classes. This is due to the great variety of objects belonging to this class, since essentially it serves as quasi-class \textit{Other}. Performance evaluation of classifiers for such fine-grained elements of fa\c{c}ade furniture and urban furniture could hardly be evaluated in the past due to the mostly insufficient representation of these fine structures resulting from the poor resolution of former datasets. Therefore this is a unique feature of H3D. 

In this context, a significant amount of misclassification can be observed for cars, which are predicted as \textit{Urban Furniture}. This can be explained by a great variety of colors in both classes and some special cases, such as the car-like shapes of covered wood piles often found in gardens (see the results in Figure~\ref{fig:predPC} (c) and (d) and their associated confusion matrices in (e) and (f)). The classes \textit{Vertical Surface} and \textit{Chimney} newly introduced compared to V3D, seem to be demanding as well. In this context, it is worth mentioning that whereas the RF classifier confuses fa\c{c}ades with other vertical surfaces as anticipated, this is not the case for the SCN, where only a small number of vertical surface points are allocated in class \textit{Fa\c{c}ade}. This may be due to the larger receptive field of \unit[20]{m} for the lowest SCN filters compared to the maximum neighborhood radius for feature computation of \unit[5]{m} in case of the RF. For chimneys, on the other hand, the RF classifier performs better probably due to the discretization of the SCN approach, so that the RF might capture such small structures more precisely for its pointwise working principle.

To conclude, our baseline solutions indicate, that the depiction of detailed structures and the consequently expanded class catalog of H3D (compared to V3D) poses new challenges for the development of methods for semantic segmentation. Particularly, this applies to entities belonging to newly introduced classes but also for the interaction of those with representatives of common object classes (e.g., \textit{Vertical Surface} vs. \textit{Fa\c{c}ade}).

\subsubsection{H3D(Mesh)}\label{sec:discussionMesh}

\begin{figure}[htbp]
	\begin{center}
		\subfigure[Labels predicted by RF]{\includegraphics[width=0.49\columnwidth,trim=10 10 10 10, clip]{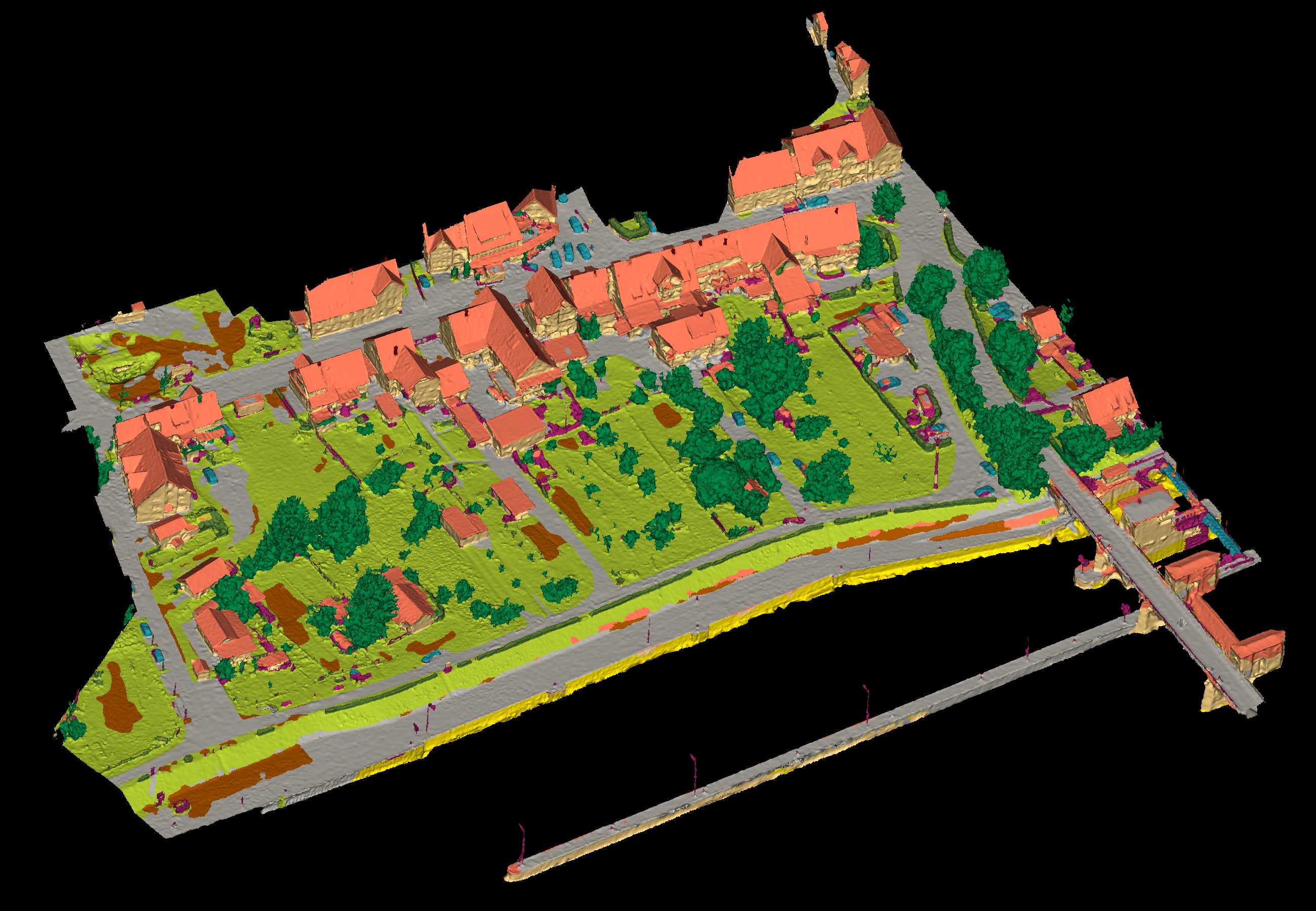}}~ 
		\subfigure[Labels predicted by SCN]{\includegraphics[width=0.49\columnwidth,trim=10 10 10 10, clip]{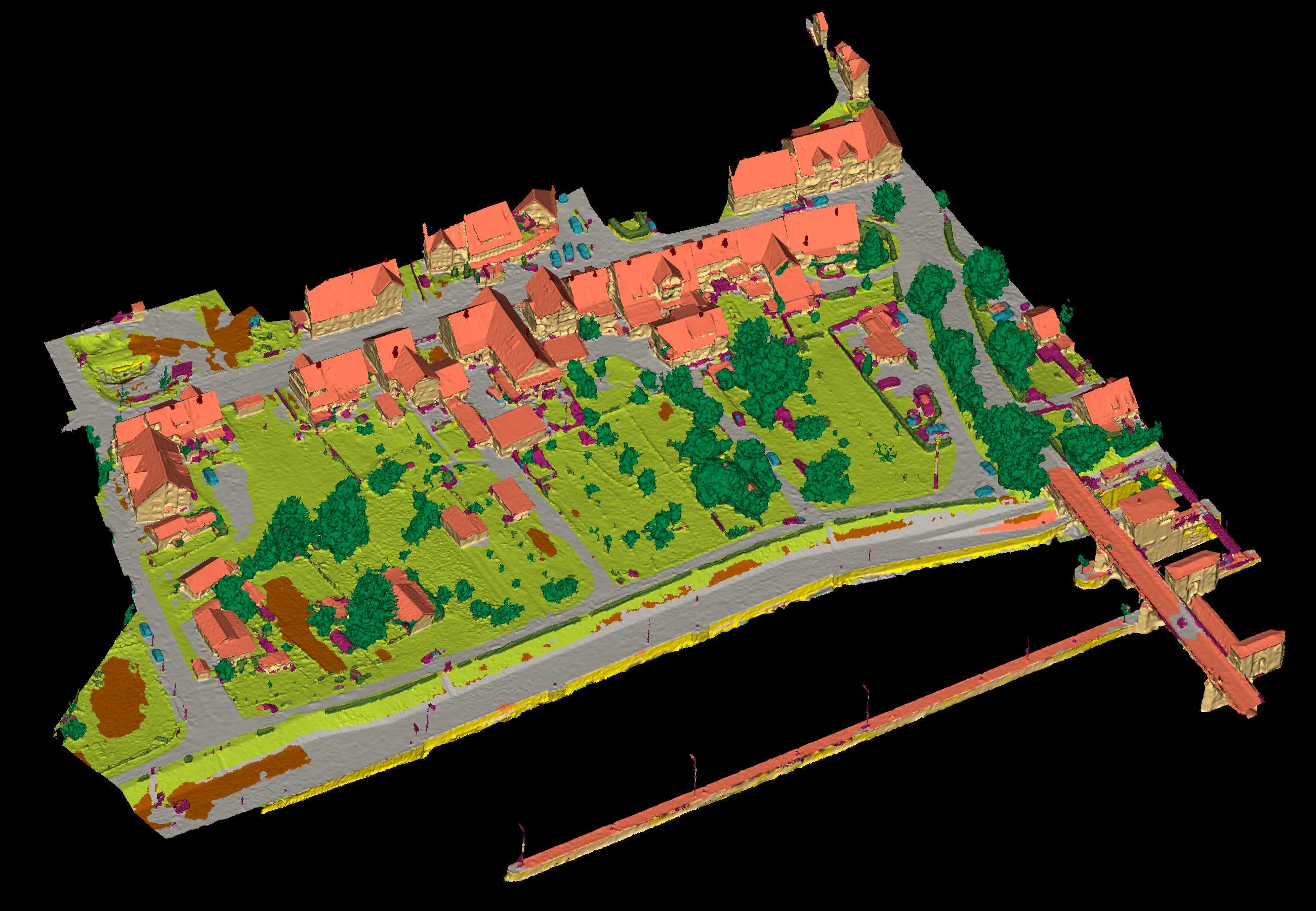}}~ \\
		\subfigure[Labels predicted by RF - close up]{\includegraphics[width=0.49\columnwidth,trim=10 10 10 10, clip]{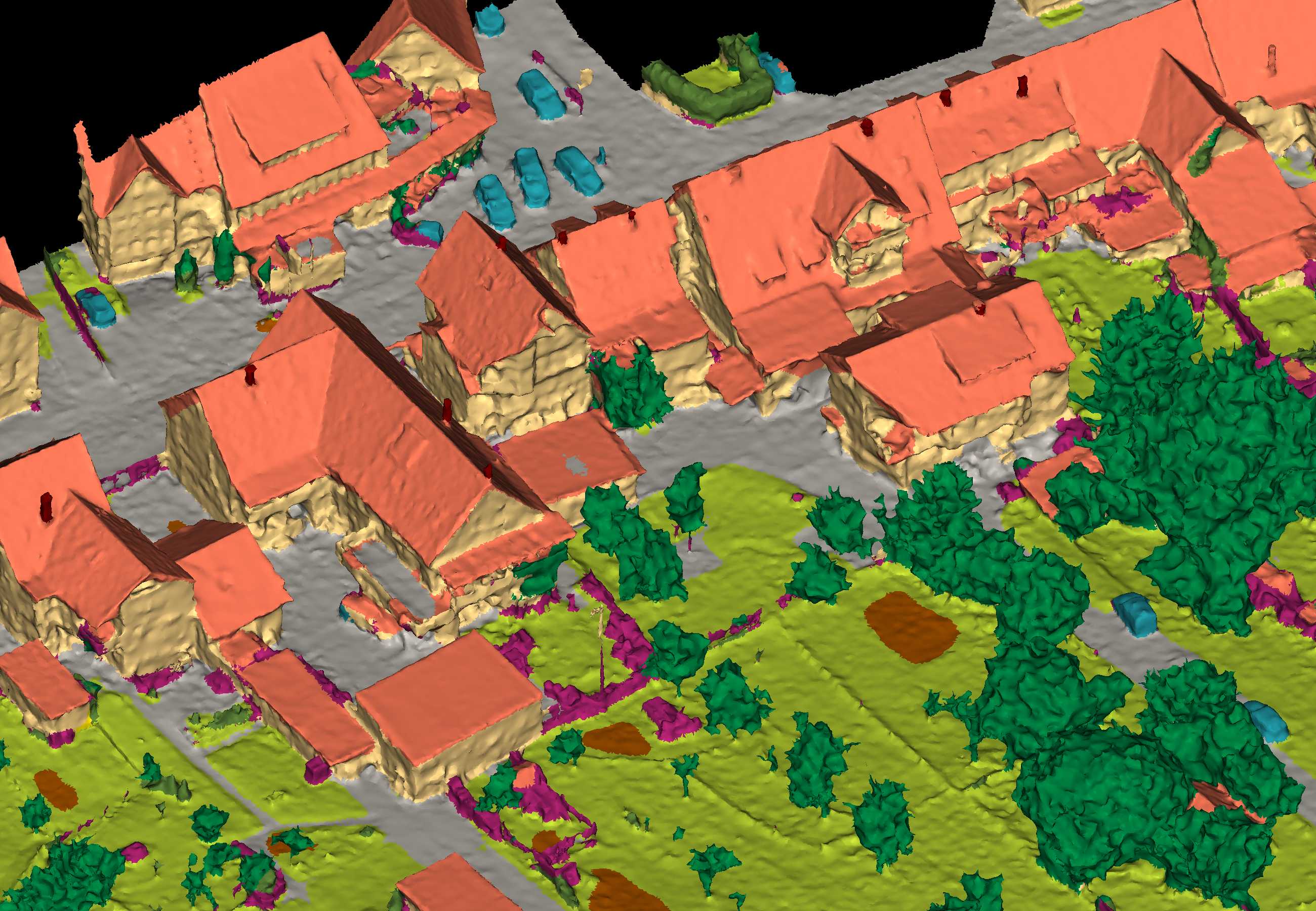}}~ 
		\subfigure[Labels predicted by SCN - close up]{\includegraphics[width=0.49\columnwidth,trim=10 10 10 10, clip]{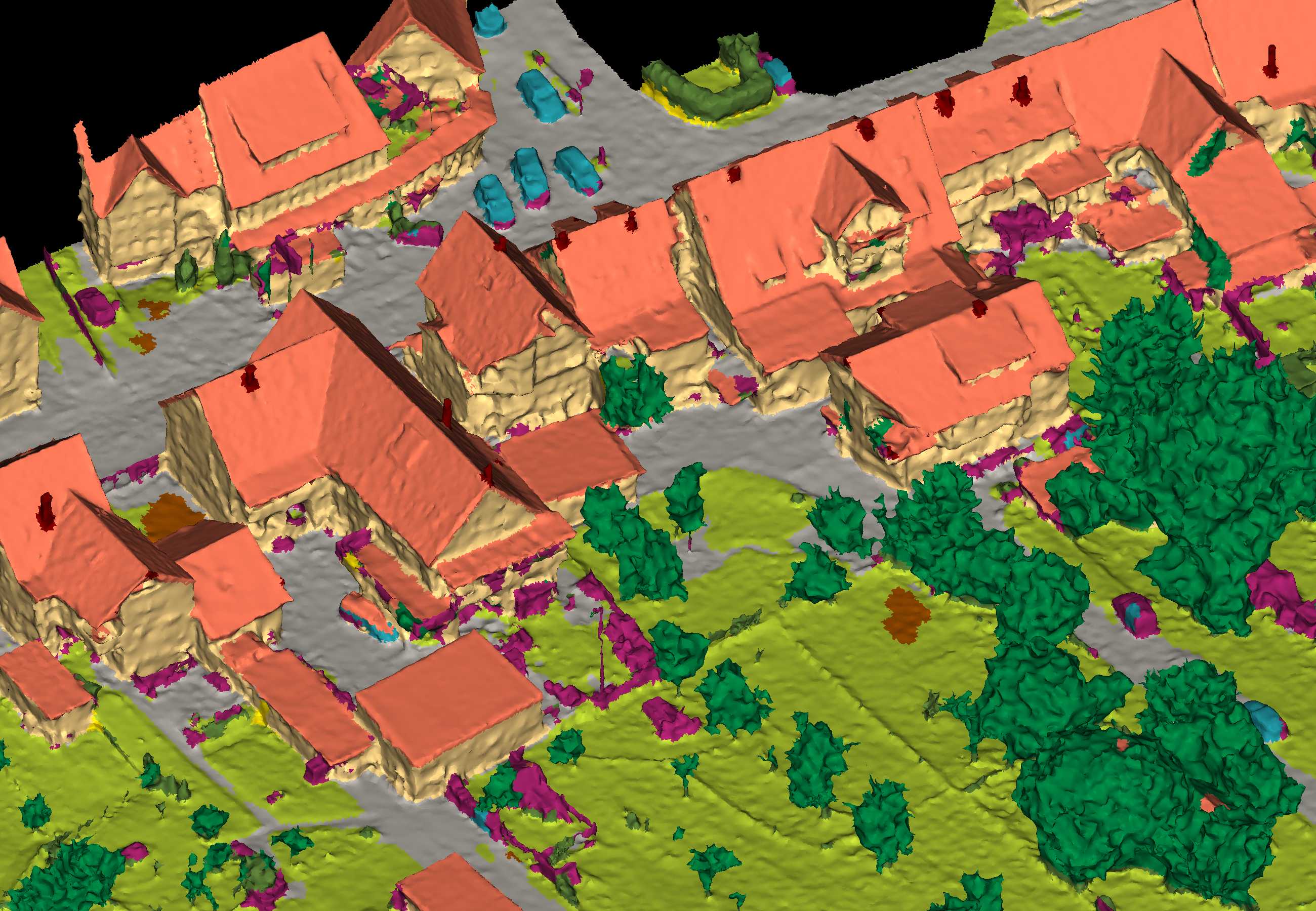}}~ \\
		\subfigure[Normalized confusion matrix for the RF classifier (weighted by face area)]{\includegraphics[width=0.49\columnwidth,trim=0 0 0 22, clip]{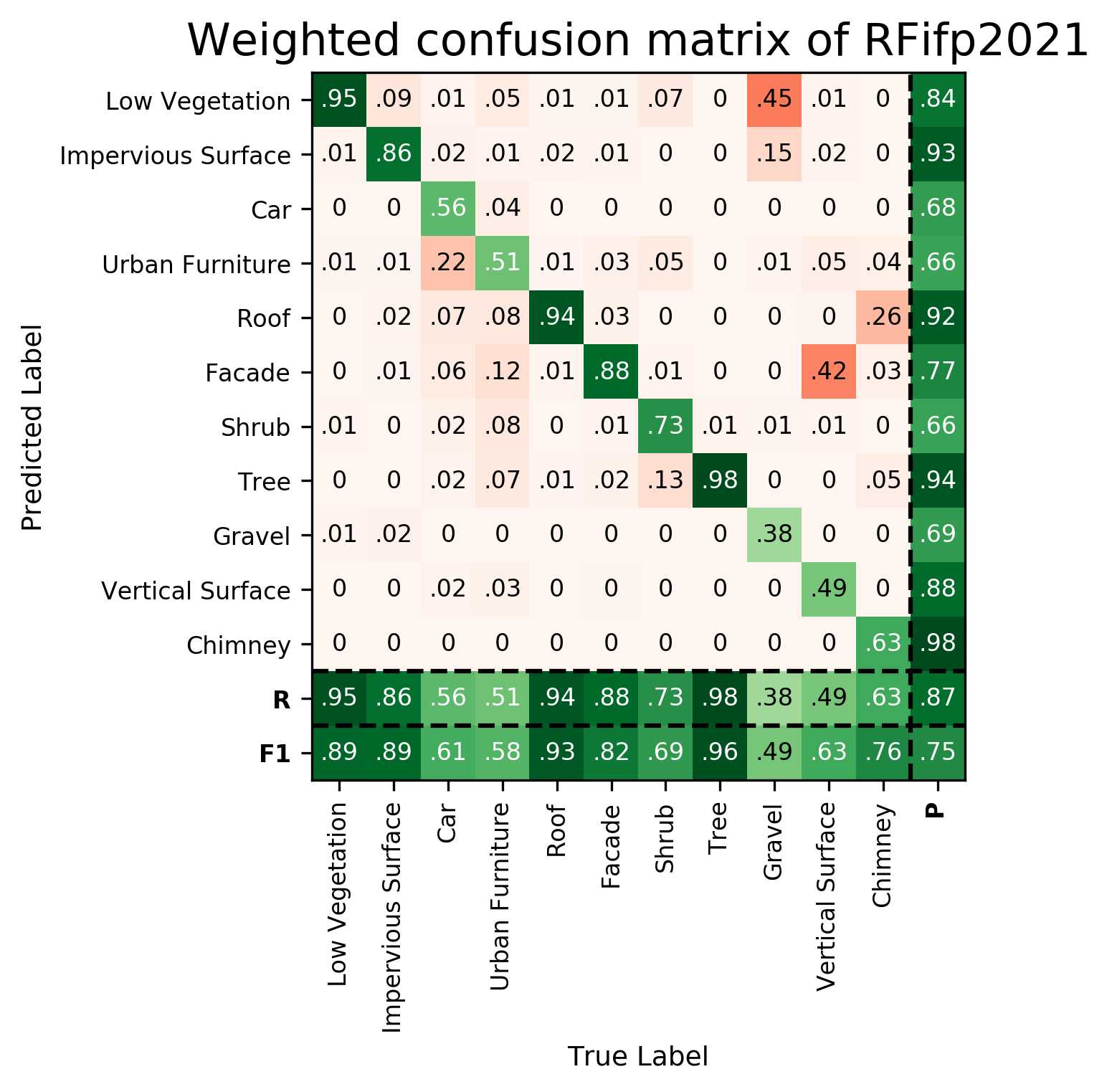}}~ 
		\subfigure[Normalized confusion matrix for the SCN classifier (weighted by face area)]{\includegraphics[width=0.49\columnwidth,trim=0 0 10 22, clip]{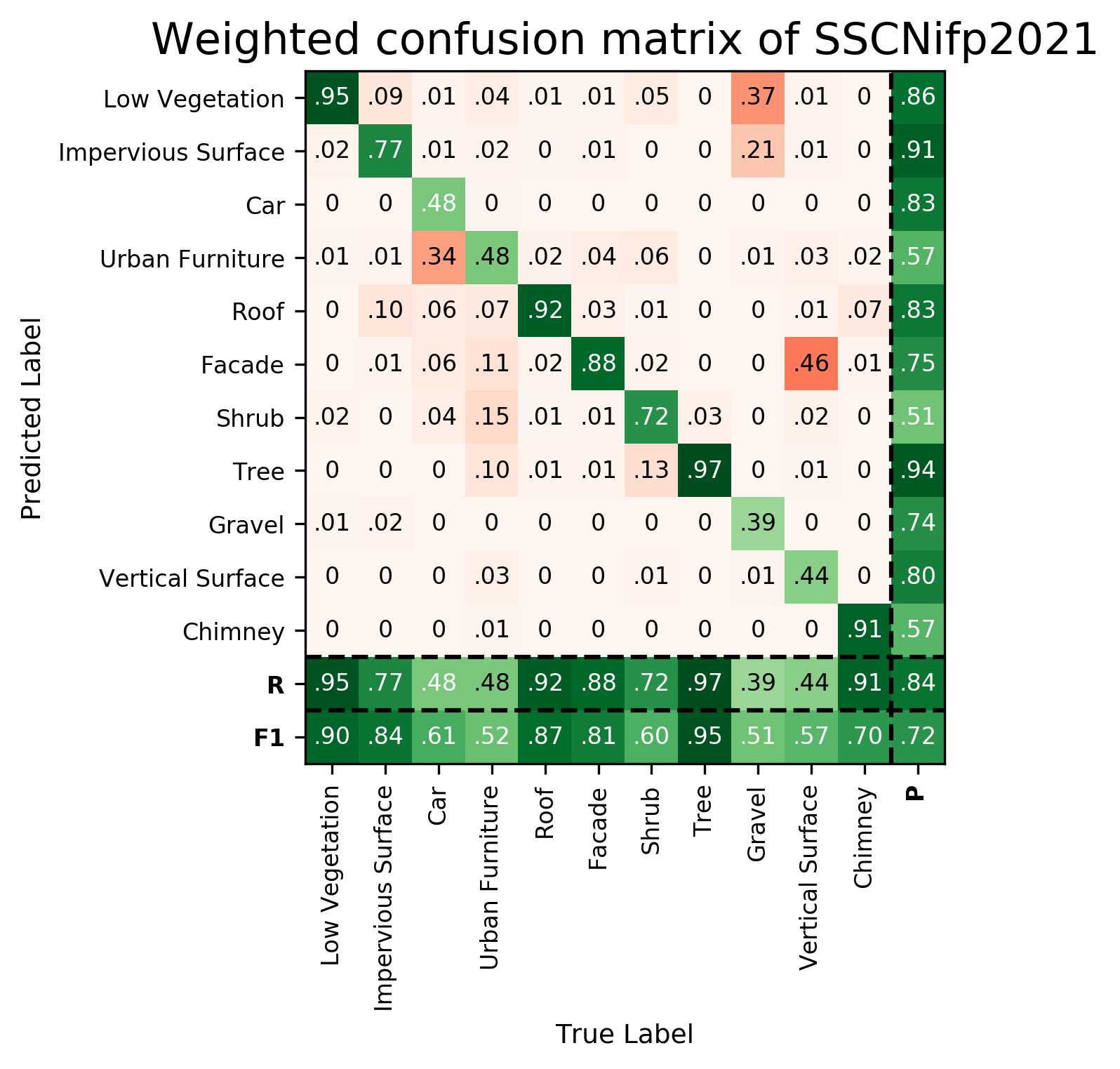}}~ \\		\caption{Results of semantic segmentation on the test set of H3D(Mesh) for both our RF (\textit{left}) and our SCN classifier (\textit{right}).}
		\label{fig:predMesh}
	\end{center}
\end{figure}
In addition to results for H3D(PC), Table~\ref{tab:baseline} also reports per-class F1-scores, mF1, and OA for our baseline classifiers RF and SCN on H3D(Mesh). The respective confusion matrices are depicted in Figure~\ref{fig:predMesh} (e) and (f). 
Due to the non-uniformity of meshes, we evaluate the considered metrics concerning the covered surface for each entity. 
Each face contributes to the performance metrics depending on its surface area. 
By these means, a large face has more impact compared to a small face.
This contrasts with point cloud evaluation where all points share the same weight. 
However, we observed that the weighted performance metrics scarcely differ from their unweighted counterparts, which indicates that the majority of large faces is correctly predicted.
Moreover, their similarity hints at the constitution of the mesh.
The automatically generated mesh differs from an ideal mesh in such sense that the non-uniformity of faces is kept small in order to uphold details in the reconstructed mesh.
Loosely speaking, a large number of faces roughly share the same face area. 
Nonetheless, flat surfaces are represented by few large faces, whereas rough surfaces (e.g., vegetational classes) use many small faces (covering roughly the same area). 
Therefore, the effect of the surface-driven evaluation is best visible for classes that consist of flat surfaces. For instance, \unit[42]{\%} of faces predicted as vertical surfaces are fa\c{c}ades in reality (for SCN).
Regarding covered area, this makes already \unit[46]{\%} of mispredicted facades. 
Generally, the feature-driven RF outperforms the data-driven SCN in terms of OA and mF1 by $2.8$ and $3.5$ percent points respectively. 
We assume that voxelization and the small set of input features mainly cause the slight inferiority of the SCN. 
Whereas the RF uses the underlying mesh geometry encoded in the variety of hand-crafted features (see Section~\ref{sec:RF}), SCN first voxelizes the textured data and learns features merely on the voxel representation. 
Furthermore, by utilizing transferred LiDAR features, RF implicitly leverages the fine-grained point cloud geometry that does not suffer from mesh reconstruction errors. 
Generally, fine-grained structures such as urban furniture and shrubs are demanding objects for meshing algorithms.
Hence, the transferred LiDAR features enhance the geometric information and help to correctly classify non-correctly reconstructed structures. 
For these reasons, we deduce that the end-to-end learning approach cannot fully compete with the feature-driven RF in this case. 
The results further indicate that feature engineering and multi-modal feature transfer are valid alternatives to state-of-the-art end-to-end learning approaches. 
Apart from their varying overall performance, the confusion matrices of RF and SCN show similar strengths and weaknesses. 
Considering per-class F1-scores, we note similar performances for classes \textit{Low Vegetation}, \textit{Vehicle}, \textit{Fa\c{c}ade} and \textit{Tree}. 
Both classifiers struggle with class \textit{Urban Furniture} due to its large intra-class variance (see discussion in Section~\ref{sec:discussionPC}). 
For instance, cars are often classified as \textit{Urban Furniture} by mistake. However, due to the utilized superior geometric information, the RF copes better with the variance resulting in an F1-score that is $5.4$ percent points better compared to SCN. 
For class \textit{Shrub}, the RF is 9.5 percent points better. In case of the SCN, the majority of predictions of class \textit{Shrub} truly belongs to \textit{Urban Furniture}. 
As can be seen for the shipping lock in Figure~\ref{fig:predMesh} (b), SCN confuses \textit{Impervious Surface} with \textit{Roof} and hence performs worse than the RF on these classes. 
The geometric similarity of ground classes in the mesh representation (\textit{Impervious Surface}, \textit{Low Vegetation}, and \textit{Soil Gravel}) makes it demanding to correctly separate them. 
Therefore, the color information is decisive for the correct prediction. 
Both classifiers predict other ground classes for \textit{Gravel/Soil}. 
\textit{Soil/Gravel} is the only class where SCN outperforms the RF. 
This indicates that SCN mostly learns geometric features. 
Most probably, the Gaussian smoothed features cause misprediction of chimneys as \textit{Roof} for RF. 
In case of SCN, \textit{Chimney} has a high recall but at cost of good precision. 
On the contrary, RF has a significantly worse recall but very high precision resulting in a better F1-score. 
The distinction of vertical surfaces and fa\c{c}ades is demanding for both classifiers due to their small inter-class variance.

\section{Conclusion}
\label{sec:conclusion}
In this paper, we presented a new benchmark on semantic segmentation of high-resolution 3D~point~clouds and textured meshes as acquired and derived from UAV LiDAR and Multi-View-Stereo: Hessigheim~3D~(H3D). We have introduced the multi-modal data corresponding to the first epoch of H3D (March 2018), comprising H3D(PC) and H3D(Mesh). Follow-on epochs (November 2018 \& March 2019) will cover an extended area and offer an even more detailed class catalog. Apart from discussing the data acquisition and processing, we applied different classifiers to H3D(PC) and H3D(Mesh) as a baseline for the benchmark. The results indicate great potential for testing ML approaches on H3D due to its large sets of labeled data.  Eventually, we hope H3D to become a second established ISPRS benchmark dataset in company with V3D.

\section*{Acknowledgements}\label{ACKNOWLEDGEMENTS}
The H3D dataset has been captured in the context of an ongoing research project funded by the German Federal Institute of Hydrology (BfG). We would like to thank the University of Innsbruck for carrying out the flight missions. Our gratitude goes to Markus Englich for providing and maintaining H3D's IT infrastructure. We appreciate the funding of H3D as an ISPRS scientific initiative 2021 and the financial support of EuroSDR. The authors would like to show their gratitude to the State Office for Spatial Information and Land Development Baden-Wuerttemberg for providing the ALS point clouds of the village of Hessigheim.

\bibliography{main}

\end{document}